\documentclass[acmtog,authorversion]{acmartarxiv}

\setcitestyle{square}

\usepackage[ruled]{algorithm2e} 

\SetAlFnt{\small}
\SetAlCapFnt{\small}
\SetAlCapNameFnt{\small}
\SetAlCapHSkip{0pt}
\IncMargin{-\parindent}

\setcopyright{rightsretained}

\usepackage{tabularx}
\usepackage{color}
\usepackage{amssymb}
\usepackage{listings}
\usepackage{textcomp}
\usepackage{cancel}
\usepackage{enumitem}
\usepackage[normalem]{ulem}
\usepackage{bm}
\usepackage{subfig}
\usepackage{multirow}
\usepackage{caption}
\usepackage{textcomp}
\usepackage[percent]{overpic}

\definecolor{lightgray}{rgb}{0.8,0.8,0.8}
\definecolor{darkblue}{rgb}{0,0,0.7}
\definecolor{darkgreen}{rgb}{0,0.5,0}
\definecolor{darkpurple}{rgb}{0.5,0.0,0.4}
\definecolor{midred}{rgb}{0.8,0,0}
\definecolor{midgreen}{rgb}{0,0.8,0}
\definecolor{midorange}{rgb}{0.8,0.6,0}

\newcommand{\new}[1]{\textcolor{darkblue}{#1}}

\renewcommand{\new}[1]{#1}

\newcommand{\nonarXiv}[1]{#1}
\renewcommand{\nonarXiv}[1]{}

\title{Snapshot Difference Imaging using Time-of-Flight Sensors}

\author{Clara Callenberg}
\affiliation{%
  \institution{Institute for Computer Science II, University of Bonn}
  \city{Bonn}
  \country{Germany}}
  \author{Felix Heide}
\affiliation{%
  \institution{Stanford University}
  \city{Stanford}
  \country{USA}}
  \author{Gordon Wetzstein}
\affiliation{%
  \institution{Stanford University}
  \city{Stanford}
  \country{USA}}
  \author{Matthias Hullin}
\affiliation{%
  \institution{Institute for Computer Science II, University of Bonn}
  \city{Bonn}
  \country{Germany}}


\begin{document}
\begin{teaserfigure}
\centering
\includegraphics[width=\textwidth]{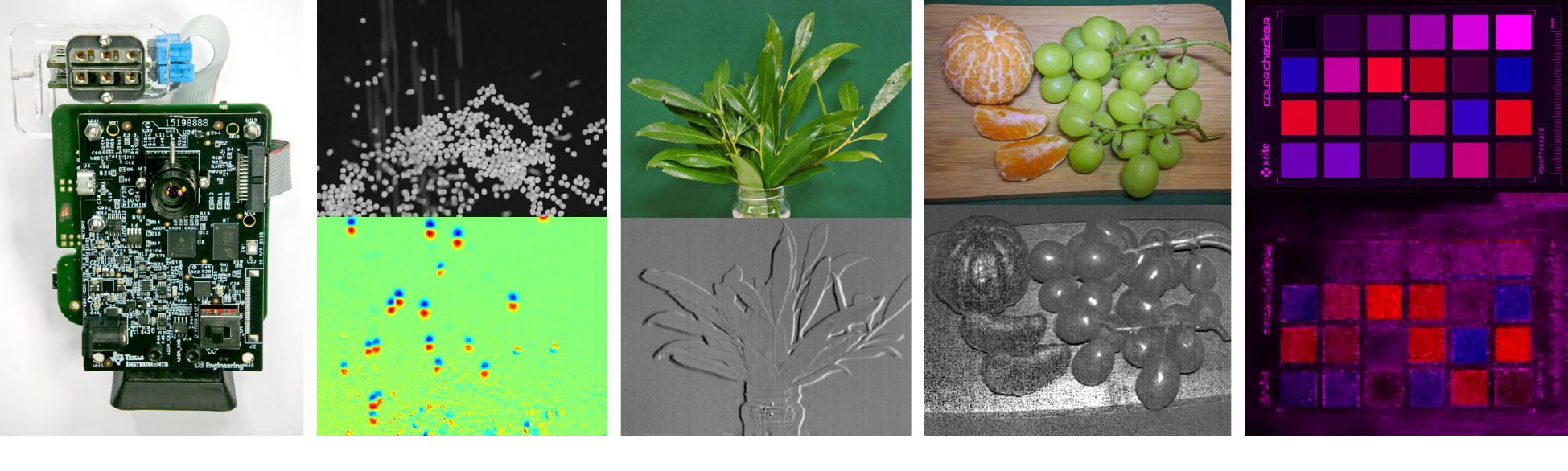}\\[-4mm]
\hfill(a)\hfill\hfill(b)\hfill\hfill(c)\hfill\hfill(d)\hfill\hfill\hspace{0.8em}(e)\hfill\,
\vspace{-2mm}
   \caption{We introduce a method that uses time-of-flight (ToF) imagers not for measuring scene depth, but rather as analog computational imagers that can directly measure difference values at a pixel level. We demonstrate this principle with a slightly modified ToF camera system (a), where simple reconfigurations in the optical path enable a wide range of imaging modalities. For instance, our system can directly sense temporal gradients (b), depth edges (c), direct-light-only images (d) and spatial gradients (Fig.~\ref{fig:crystal_edges}) -- each in a single exposure and without any additional decoding steps. We further show that the remarkable noise statistics of such imagers can be exploited to extract two color channels (here: red and blue) from a single snapshot taken under red \emph{and} blue illumination (e). The top images in columns (b)--(e) are reference photographs of the respective scenes; the bottom ones visualize the output of our system.}
   \label{fig:teaser}
\end{teaserfigure}

\begin{abstract}
	Computational photography encompasses a diversity of imaging techniques, but one of the core operations performed by many of them is to compute image differences. An intuitive approach to computing such differences is to capture several images sequentially and then process them jointly. Usually, this approach leads to artifacts when recording dynamic scenes. In this paper, we introduce a snapshot difference imaging approach that is directly implemented in the sensor hardware of emerging time-of-flight cameras. With a variety of examples, we demonstrate that the proposed snapshot difference imaging technique is useful for direct-global illumination separation, for direct imaging of spatial and temporal image gradients, for direct depth edge imaging, and more.
\end{abstract}

\thanks{This work was supported by the German Research Foundation (HU-2273/2-1) and the X-Rite Chair for Digital Material Appearance.}

\maketitle

\section{Introduction and motivation}
\label{sec:intro}

Over the last two decades, research in computational photography has been striving to overcome limitations of conventional imagers via a co-design of optics, sensors, algorithms, and illumination. Using this paradigm, unprecedented imaging modalities have been unlocked, such as direct-global light transport separation~\cite{nayar2006fast}, gradient imaging~\cite{tumblin2005gradient}, temporal contrast imaging~\cite{lichtsteiner2008temporalcontrast}, and direct depth edge imaging via multi-flash photography~\cite{Raskar:2004}. A common operation for many of these techniques is to record two or more images and then compute the difference between them. Unfortunately, difference imaging is challenging for dynamic scenes, because motion creates misalignment between successively captured photographs which is in many cases difficult to mitigate in post-processing. In this paper, we explore a new approach to capturing difference images in a single exposure and generalize difference imaging to a variety of applications.

\begin{figure}[t]
\centering
\includegraphics[width=0.8\columnwidth]{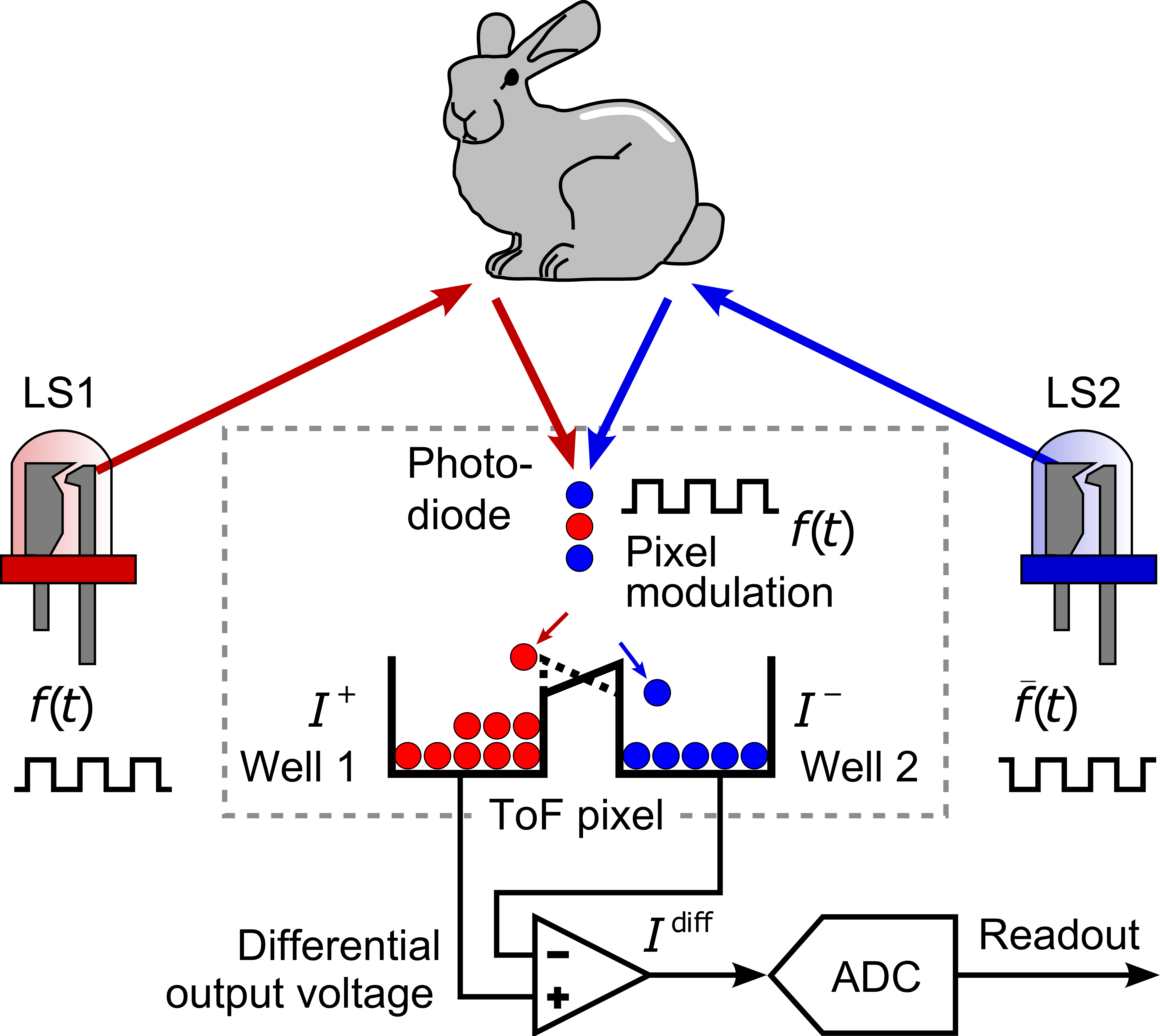} 
\caption{Principle of operation of a time-of-flight (ToF) pixel. A light source is temporally modulated, its emitted light is reflected by the scene, and then demodulated in the pixel. To demodulate the coded illumination in the detector, two wells in each pixel collect charge carriers and an electric field oscillates at the demodulation frequency to direct incident photoelectrons into one or the other well. The sensor circuit measures the voltage difference before digitizing it by an analog-to-digital converter (ADC). Here, we illustrate how the difference between two modulated light sources can be directly measured with such a pixel architecture.}
\label{fig:pixel}
\end{figure}

We propose to re-purpose time-of-flight (ToF) sensors to facilitate instantaneous difference imaging. The usual application for these sensors is depth imaging. In that context, they are operated in conjunction with a periodically modulated light source. Light that has been reflected by the scene is demodulated by the sensor, reconstructing the shift in modulation phase and thereby the depth estimate per pixel. This functionality is achieved by a pixel architecture that employs two potential wells for photoelectrons to be stored in during the exposure, and that subtracts the charges accumulated in these two wells (Fig.~\ref{fig:pixel}). In other words, the core functionality of time-of-flight sensors is based on being able to take the difference of two incident signals before analog-to-digital (A/D) conversion.

Rather than computing scene depth, we demonstrate how ToF sensing technology can be used to conveniently implement a range of computational photography techniques, including direct-global separation, direct depth edge imaging, spatio-temporal gradient imaging, and more. The capabilities unlocked with snapshot difference imaging are particularly interesting for applications that require low-power, low-latency on-board processing with low bandwidth communication channels, such as internet-of-things devices. With this paper, we take first steps towards these directions.

Specifically, our contributions are the following:
\begin{itemize}[leftmargin=*]%
\setlength\itemsep{0.2em}
	\item We introduce the concept of generalized difference imaging and develop an image formation and a noise model for this principle.
	\item We construct a prototype difference imager using a modified time-of-flight camera combined with multiple, spatio-temporally coded light sources.
	\item We evaluate the proposed imaging concept with several practical applications, including direct-global separation, direct depth edge as well as spatio-temporal gradient imaging. 
	\item We demonstrate that two images can be recovered from a single difference image by exploiting characteristics of the proposed image formation model. 
\end{itemize}

\paragraph{Overview of benefits and limitations}

\new{The proposed method has two primary benefits. First, capturing a difference image within a single exposure allows for faster time scales to be recorded than capturing two separate images and subtracting them digitally. Second, the noise properties of difference imaging before A/D conversion are shown to be favorable over digital subtraction post A/D conversion. A limitation of the proposed technique is that it relies on ToF sensors, which currently provide much lower resolution and signal quality than well-established CMOS or CCD sensors. Thus, comparing digital difference imaging with CMOS sensors and analog difference imaging with ToF sensors may not be beneficial for the latter approach. Yet, we demonstrate that our method yields superior noise performance for sensors with comparable characteristics.}

\section{Related work}
\label{sec:relwork}

\paragraph{Computational ToF imaging} 
This work presents a method for difference imaging by re-purposing two-bucket sensors usually used for depth imaging in lock-in ToF cameras. Lock-in time-of-flight sensors are a rapidly emerging sensor technology, with Microsoft's Kinect for XBOX One as the highest-resolution sensor available on the market at $512 \times 424$ pixels~\cite{Bamji:2015}. For technical details on lock-in ToF sensors we refer the reader to~\cite{lange1999time,Hansard:2012}. A growing body of literature re-purposes these emerging sensors, in combination with computational methods, to address a variety of challenging problems in imaging and vision. Kadambi et al.~\shortcite{kadambi2013coded} reduce multi-path interference by coding the modulation profiles, Heide et al.~\shortcite{HeideSIG2013} recover temporal profiles of light transport by measuring frequency sweeps, which allows for improved imaging in scattering media~\cite{Heide2014scattering}. Recently, Tadano et al.~\shortcite{Tadano2015} design depth-selective modulation functions enabling virtual-blue screening and selective back-scatter removal as applications. 

\paragraph{Differential-pixel sensors}
The proposed difference imaging method subtracts two signals before A/D conversion by ``piggybacking'' on two-bucket ToF sensor technology. Wang et al.~\shortcite{wang2012180nm} have previously proposed a custom sensor design that also performs pre-ADC subtraction for the purpose of optoelectronic filtering and light field capture. Specifically, the authors use programmable gain operational amplifiers to compute the sum and difference of pixel pairs, which is then passed on to the A/D converter. In combination with local diffractive gratings as optics on every pixel, this allows to realize filtering with positive and negative filter coefficients. In contrast to a conventional sequential capture approach, these differential-pixel sensors offer reduced bandwidth~\cite{Wang2012}, at the cost of spatial resolution. Compared to the proposed method, the optical filters are static and prohibit the flexible modes of operation demonstrated in this work. Changing from one difference-imaging task to another would require placing a different mosaicking pattern on the sensor.

\paragraph{Differential optical systems}
Instead of this optoelectronic approach to difference imaging, one could also imagine cameras that perform the signal subtraction purely optically, plus a DC offset to ensure positivity. Building on Zomet and Nayar's work~\shortcite{zomet2006lensless}, Koppal et al.~\shortcite{Koppal2013TowardWM} present an optical design consisting of a micro-aperture mask in combination with lenslets allowing to design custom optical template filters for a variety of computer vision tasks. This approach may be adopted to design optical filters that perform spatial gradient or other filter differences in a single-shot, by designing difference filters with a DC offset to ensure non-negative coefficients. In theory, this approach would require variable high-resolution aperture patterns~\cite{zomet2006lensless}. Note also, that the proposed approach would be a natural choice for suppressing the DC in such a setup by relying on the adaptive background suppression of recent ToF sensors.

\paragraph{Event-based sensors}
A further sensor design for differential measurements are event-based dynamic vision sensors~\cite{lichtsteiner2008temporalcontrast,gottardi2009100}, which have been demonstrated for applications in vision and robotics, such as tracking~\cite{kim2016real} and simultaneous localization and mapping (SLAM)~\cite{weikersdorfer2014event}. Each pixel in such sensors asynchronously measures temporal intensity changes and generates spike events for temporal differences with magnitude above a given activation threshold. This event-driven behavior is achieved by augmenting each pixel with its self-timed switched-capacitor differencing circuit. By reading out pixels asynchronously, the core benefit of this sensor design is the the low bandwidth requirement, enabling high frame rates and low power consumption~\cite{lichtsteiner2008temporalcontrast,gottardi2009100}. However, similar to the differential-pixel sensors, this comes at the cost of reduced spatial resolution, when compared to conventional sequential capture. While the proposed solution, based on ToF sensors, shares limitations in resolution, temporal differencing sensors do not support the very flexible modes of operation shown in this work. For example, capturing intensity or depth images requires solving ill-posed inverse problems~\cite{kim2016real}.

\paragraph{Split-pixel HDR sensors}
Backside-illuminated split-pixel architectures have become the dominant choice for high-dynamic-range (HDR) imaging in high-speed vision cameras~\cite{willassen20151280}. Single-shot HDR capture is essential for vision-based autonomous or assisted driving systems where reacting to fast moving objects over a wide dynamic range is critical~\cite{solhusvik2013comparison}. A variety of HDR sensor designs for high frame rates have been proposed in the past. Skimming HDR sensors perform partial resets (draining) of the accumulated charges during integration, allowing repeated partial integration with successively shorter resets~\cite{darmont2012high}. The repeated integration can cause motion artefacts if partial saturation are reached quickly. Split-pixel architectures eliminate this issue by dividing each pixel into multiple buckets~\cite{nayar2003adaptive,wan2012cmos}. Multiple exposures are captured simultaneously by implementing different-sized photosensitive areas (OmniVision OV10640, OV10650). Given the emerge of split-pixel architectures as a key vision sensor technology, we believe that the proposed two-bucket difference imaging method may have broad applications even beyond the ones in this work.

\section{Imaging principle}\label{sec:definition}\label{sec:method}
Of all technologies that can be used for time-of-flight imaging, correlation sensors are the most widespread and affordable. This is also the type of imager we are using in this work; throughout the paper, we use the term ``time-of-flight (ToF) sensor'' synonymously for this particular technology. 

A pixel in a ToF sensor measures the amount of correlation between the incoming temporally varying photon flux $g_i(t)$ and a sensor modulation signal $f(t)\in[0,1]$ that also varies in time~\cite{lange1999time}. Unlike regular CCD or CMOS sensors where electrical charges generated in a photodiode are collected in a potential well, ToF sensors feature two such wells per pixel (Fig.~\ref{fig:pixel}). The sensor modulation $f(t)$ decides whether a charge generated at time $t$ will tend to end up in one well or the other. At the end of the integration phase, the difference between the two wells is read out and digitized. Neglecting quantization from the A/D conversion, this results in the digital value
\begin{equation}
\hat I^\text{diff} = \rho\cdot\eta\cdot(\hat I^+-\hat I^-),
\label{eq:pixeleff}
\end{equation}
where $\rho$ is the conversion factor from electron counts to digital units, and $\eta$ denotes the so-called \emph{demodulation contrast} \cite{schmidt2011analysis}. 
$\hat I^+$ and $\hat I^-$ are the photoelectrons collected in the two wells over the integration period $[0,T]$:
\begin{equation}
{\hat I^+\brack \hat I^-}=\int_0^T{f(t)\brack 1-f(t)}g_i(t)dt
\label{eq:wellmeasurement}
\end{equation}
The incoming photon rate $g_i(t)$ is a function of the scene and the time-varying intensity $g(t)$ of an active light source that illuminates it. In ToF imaging, $f(t)$ and $g(t)$ are periodic functions of the same high frequency, typically 20--100\,MHz, and the delay of light propagating from a source to the sensor results in a relative phase shift which is measured to recover depth. 
In snapshot difference imaging, we introduce two modifications to this scheme. Firstly, we reduce the modulation frequency to a point (1--5\,MHz) where the propagation of light through near-range scenes can be assumed to be instantaneous and $f(t)$, typically generated by a digital circuit, only assumes the values 0 and 1. Secondly, we use two light sources, one (LS1) driven using the same function $f(t)$ and the other one (LS2) with its logical negation $\bar f(t)$. According to Eq.~\ref{eq:wellmeasurement}, the photocharges collected in $\hat I^+$ will record an image of the scene as illuminated by LS1, and LS2 will fill $\hat I^-$. The pixel value $\hat I^\text{diff}$ thus measures the difference between two images taken under different illumination conditions, an insight that forms the foundation of this work.

\subsection{Noise model}
Time-of-flight imagers are complex photonic devices and as such suffer from noise of various different sources \cite{schmidt2011analysis}. The differential measurement scheme, and in particular the multi-tap measurement schemes typically used in ToF operation, cancel out many of the systematic errors introduced by the hardware. None of these measures, however, are capable of removing shot noise, which is the uncertainty that occurs during the counting of photoelectrons.

If $\hat I^\pm$ are the \emph{expected} electron counts for the two wells, the \emph{actual} number of collected electrons $I^\pm$ in any image recorded is a Poisson-distributed random variable with mean $\mu^\pm$ and variance $(\sigma^\pm)^2$ that are both identical to the respective expected value:
\begin{equation}
\mu^\pm=(\sigma^\pm)^2=\hat I^\pm
\label{eq:poissonstats}
\end{equation}
As the difference of two independent random variables, the final pixel value is also a random variable, and it follows a Skellam distribution \cite{skellam,hwang2012difference}.
Mean $\mu_\textrm{diff}$ and variance $\sigma^2_\textrm{diff}$  relate to the means $\mu_\pm$ and variances $\sigma_\pm^2$ of $I_\pm$ as
\begin{eqnarray}
\textstyle	\mu_\textrm{diff}=&\eta\left(\mu_+ - \mu_-\right) &= \eta\left(\hat I_+ - \hat I_-\right)\\ 
\textstyle	\sigma_\textrm{diff}^2=&\eta^2\left(\sigma_+^2 + \sigma_-^2\right)+\sigma^2_\textrm{read} &= \eta^2\left(\hat I_+ + \hat I_-\right)+\sigma^2_\textrm{read}
\label{eq:skellamvar}
\end{eqnarray}
where $\sigma^2_\textrm{read}$ models additional noise sources (assumed to be zero-mean), and the device constant $\eta\!\in\![0,1]$ is the imager's contrast 
 \cite{schmidt2011analysis}. In terms of a matrix-vector product:
\begin{equation}
\begin{pmatrix}
\mu_\textrm{diff} \\ \sigma_\textrm{diff}^2 - \sigma^2_\textrm{read}
\end{pmatrix}
= 
\underbrace{
\begin{pmatrix}
\eta & -\eta \\ \eta^2 & \eta^2
\end{pmatrix}}_H
\begin{pmatrix}
\hat I^+ \\ \hat I^-
\end{pmatrix}.
\label{eq:skellammatrix}
\end{equation}
Note that the uncertainty $\sigma_\textrm{diff}^2$ of the measurement $I^\text{diff}$ depends not primarily on the net difference value, but rather on the latent components $\hat I^\pm$ that are subtracted from one another. Thus, even when there is zero signal ($\hat I^\text{diff}=0$), the actual observation $I^\text{diff}$ can suffer from significant noise. This is a principal property of difference imaging and holds for all sorts of settings, applying to in- as well as post-capture differencing techniques. 
We call the system-specific matrix $H$ the \emph{Skellam mixing matrix}. In Section~\ref{sec:separation}, we show how it can be used to recover two color channels from a single exposure --- an insight that, to our knowledge, has not been discovered before.

\section{Prototype difference imager}
\newcommand{\overbar}[1]{\mkern 1.5mu\overline{\mkern-1.5mu#1\mkern-1.5mu}\mkern 1.5mu}

\begin{figure}[tbp]%
\centering
\includegraphics[width=0.8\columnwidth]{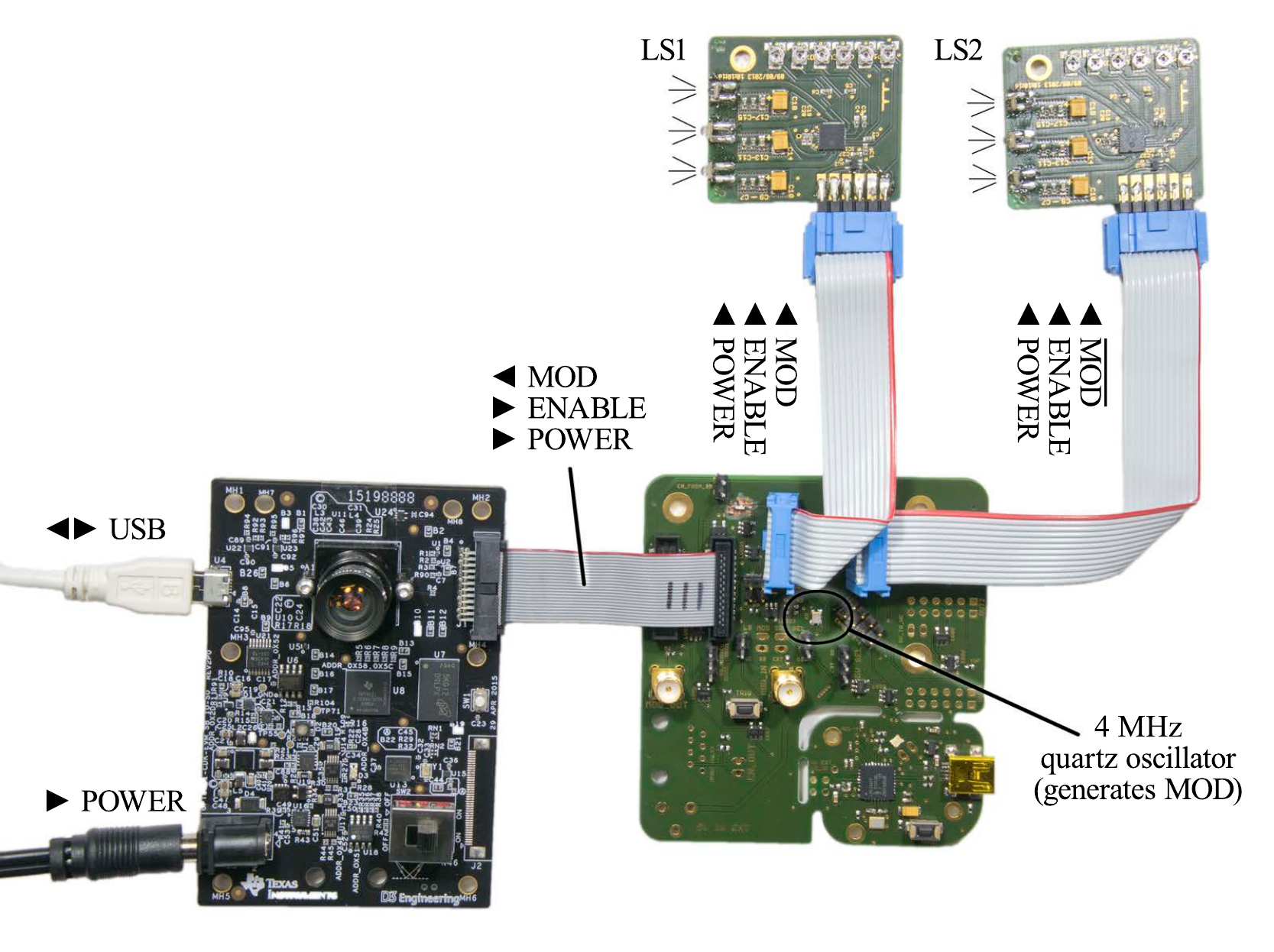}
\vspace{-3mm}
\caption{Components of our system based on the Texas Instruments OPT8241-CDK module (bottom left). An external function generator (bottom right) produces a relatively low-frequency square wave, $\textrm{MOD}$, and its negated version, $\overbar{\textrm{MOD}}$, that correspond to $f(t)$ and $\bar f(t)$ and modulate the sensor and two light sources (top right). The ENABLE signal is held high by the camera during  integration.}%
\label{fig:setup}%
\end{figure}

We constructed snapshot difference imagers based on two different time-of-flight sensing platforms. Our first prototype (not pictured) is a recreation of Heide et al.'s system that is based on the discontinued PMD Technologies CamBoard nano~\shortcite{HeideSIG2013}. The second prototype (Fig.~\ref{fig:teaser} and~\ref{fig:setup}) combines the Texas Instruments (TI) OPT8241-CDK evaluation module with external modulation and light sources in a similar way to the system described by Shrestha et al.~\shortcite{Shrestha:2016}. Both imagers have their infrared-pass filters removed so they can sense visible light. (For the TI sensor, we carefully polished the filter coating off using a Dremel 462 rotary tool.) To enable difference imaging, the external light sources are configured to operate in and out of phase, respectively. Both systems are configured to capture at an exposure time of 2000\,\textmu s per frame and 60 frames per second.

Each of our light sources carries three OSRAM OSLON LEDs that are switched using the same signal. To implement the different imaging modalities described in Section~\ref{sec:applications}, we equipped the light sources with LEDs of different colors (red, blue, infrared), placed them in different positions and equipped LEDs and camera with polarization filters as required for each particular purpose.

In Appendix \ref{sec:appendix}, to lower the entry barrier for the reader, we describe  an alternative modification for stock OPT8241-CDK modules that does not require any custom hardware.

\paragraph{Measurement procedure} To reduce fixed pattern noise, a black frame was recorded before the data acquisition with our setup, and later subtracted from each measured frame. As a result, the difference image pixels obtain negative and positive values, depending on the charge balance of the two potential wells. This applies to all results shown in this paper.
\section{Applications and Evaluation}
\label{sec:applications}

\subsection{Polarization-based direct-global separation}
\label{sec:direct_global}

Many computational imaging techniques are based on a decomposition of light transport into a \emph{direct} component, i.e., light that has undergone exactly one scattering event between light source and camera, and a multiply scattered \emph{indirect}, or \emph{global}, component. Being able to investigate these components separately has been shown to enable, for instance, more realistic digital models for human skin \cite{ma2007rapid} or more robust 3D scans of objects made of challenging materials \cite{chen2007polarization,o2015homogeneous}.
\new{While true separation into direct and indirect components is not within reach, researchers have used common observations about light transport to derive useful heuristics: indirect light tends to be spatially low-pass filtered \cite{nayar2006fast}, it generally does not fulfill the epipolar constraint \cite{otoole2012primal} and does not preserve polarization \cite{wolff1990polarization}. Here, we use our setup to exemplarily implement the third of these heuristics in the form of a single-shot polarization-difference imager, and demonstrate its capability to isolate directly reflected light.}

\begin{figure}[tbp]
\centering
\includegraphics[width=0.75\columnwidth]{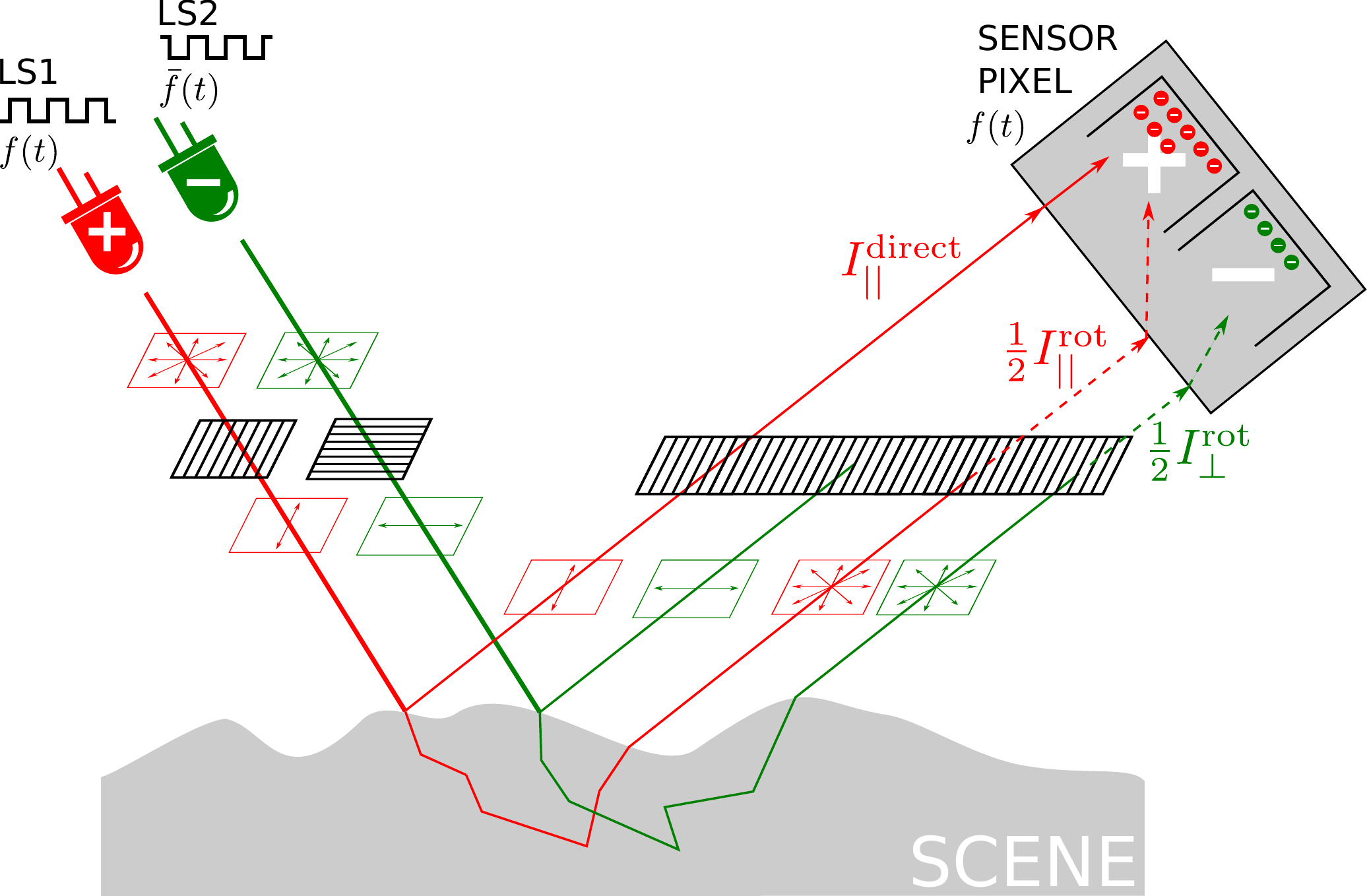}%
\vspace{-1mm}
\caption{Polarization difference imaging principle: the light of two identical light sources is modulated in ($+$) and out ($-$) of phase, respectively, with the sensor. The light sources' polarization directions are mutually perpendicular, one them being aligned in parallel with the analyzer filter in front of the sensor. Light that is reflected directly off the surface of the scene preserves the polarization while the light that scatters multiple times in the scene becomes depolarized. The sensor measures the difference image $I^\textup{diff} = I_{||}^\textup{direct} + \frac{1}{2} I_{||}^\textup{rot} - \frac{1}{2} I_{\perp}^\textup{rot} = I_{||}^\textup{direct}$.}%
\label{fig:pol_sketch}%
\end{figure}

According to Fig.~\ref{fig:pol_sketch}, illuminating the scene using two light sources with crossed linear polarization and an analyzing filter on the camera, one can consider four different components of the image: 
\begin{figure*}[t]
\centering
\begin{tabularx}{0.87\textwidth}{cccc}%
 \!\includegraphics[width = 0.21\textwidth]{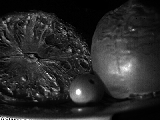}\!\!\!%
&\!\includegraphics[width = 0.21\textwidth]{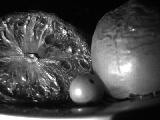}\!\!\!%
&\!\includegraphics[width = 0.21\textwidth]{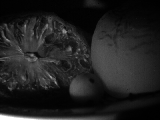}\!\!\!%
&\!\includegraphics[width = 0.21\textwidth]{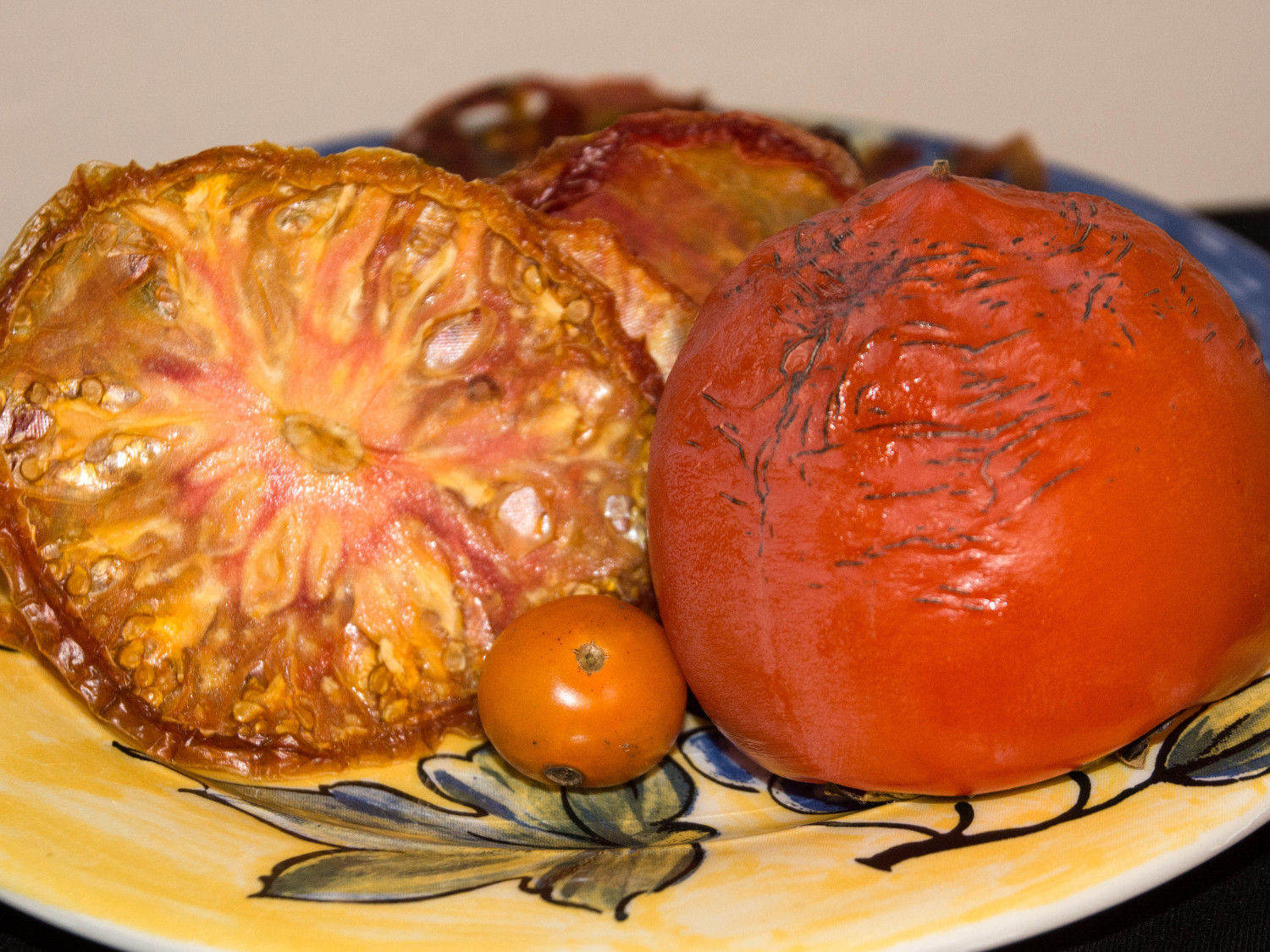}\!\!\!\\
 \!\includegraphics[width = 0.21\textwidth]{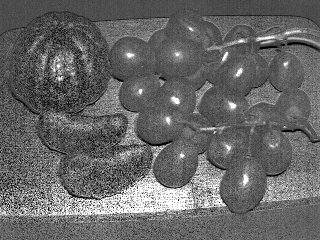}\!\!\!%
&\!\includegraphics[width = 0.21\textwidth]{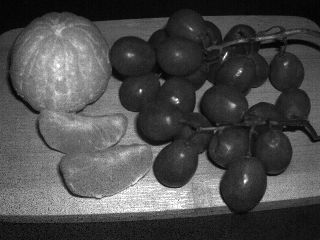}\!\!\!%
&\!\includegraphics[width = 0.21\textwidth]{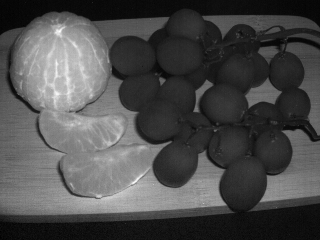}\!\!\!%
&\!\includegraphics[width = 0.21\textwidth]{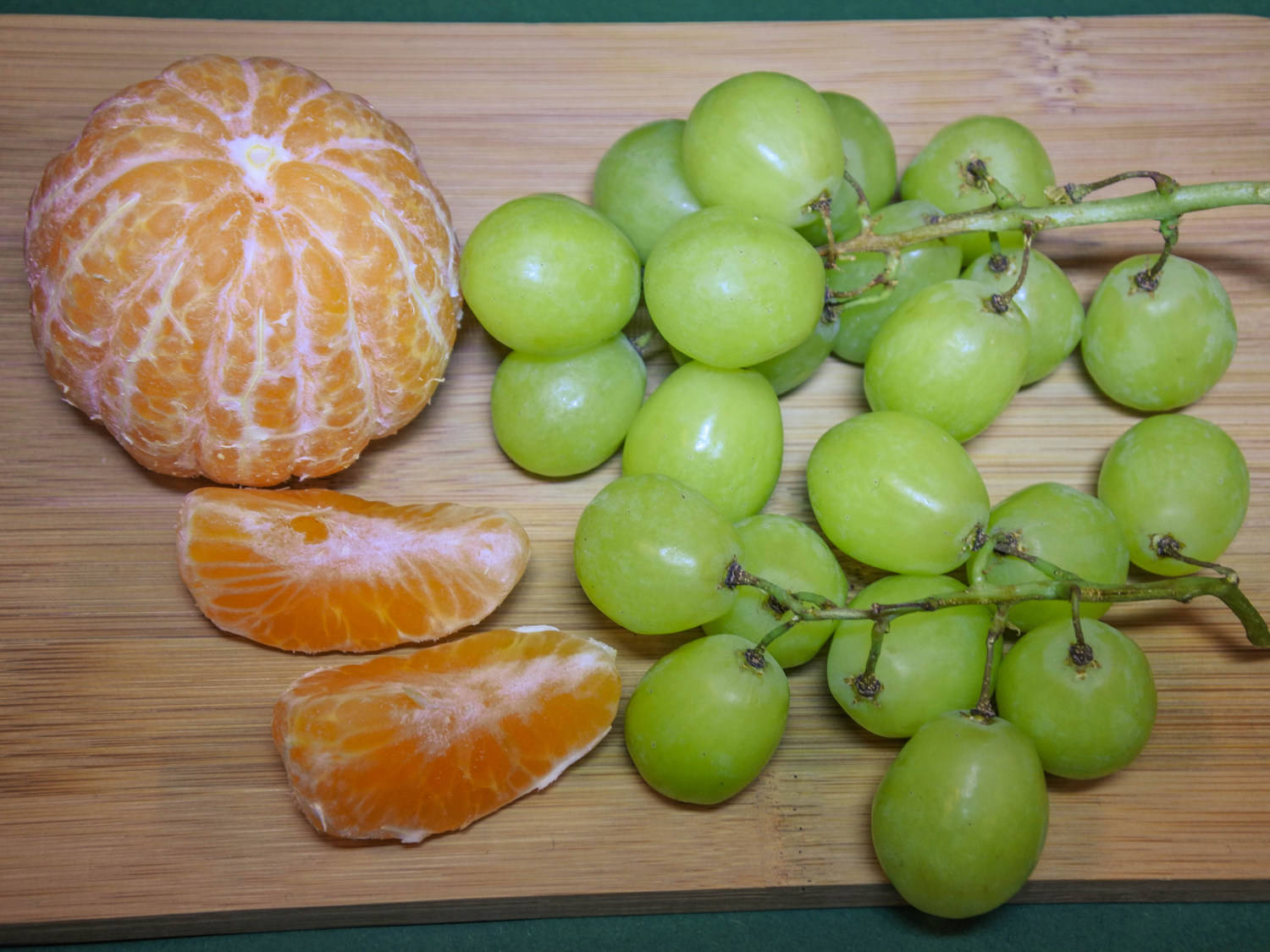}\!\!\!\\
(a)&(b)&(c)&(d)
\end{tabularx}
\vspace{-2mm}
\caption{Polarization difference images of the two different scenes seen as RGB images in column (d). Column (a) shows the difference image $I^\textup{diff}$ containing only directly reflected light. The images in column (b) depict $I^+$ (parallel polarizers), the ones in column (c) $I^-$ (crossed polarizers). The latter contains only light that has undergone multiple scattering events. The top dataset was captured with the PMD-based system, the bottom row with the TI sensor. The images have been rescaled for better visibility.}
\label{fig:polcompare}
\end{figure*}
\vspace{-1pt}
\begin{enumerate}[align=parleft, labelsep=0.1cm]
\item[$I_{||}^\text{direct}$] light that initially passed the polarization filter parallel to the analyzer and that was reflected directly in the scene, hence preserving the orientation of the polarization,
\item[$I_{||}^\text{rot}$] light that initially passed the polarization filter parallel to the analyzer and that was scattered multiple times in the scene, thus not preserving the orientation of the polarization, 
\item[$I_{\perp}^\text{direct}$] light that initially passed the polarization filter perpendicular to the analyzer and that was reflected directly in the scene, 
\item[$I_{\perp}^\text{rot}$] light that initially passed the polarization filter perpendicular to the analyzer and that was scattered multiple times in the scene. 
\end{enumerate}

Assuming that multiple scattering in the scene completely depolarizes the light for both initial directions of polarization, the amount of the light reaching the camera after illumination with light polarized in parallel with the analyzer is the component that is in phase with $f(t)$:
\[I^+ = I_{||}^\text{direct} + \frac{1}{2} I_{||}^\text{rot},\]
while the amount of light reaching the sensor after illumination with light polarized perpendicularly to the analyzer is only the ratio of the incident light that has been depolarized and therefore passes the analyzing filter:
\[I^- = \frac{1}{2} I_{\perp}^\text{rot}.\]
The light sources are identical and we assume that the ratio between directly and indirectly scattered light is equal for both initial directions of polarization, hence $I_\perp^\text{rot} = I_{||}^\text{rot}$. Therefore, the difference image that the PMD sensor acquires in lock-in operation with the light sources is 
\[I^\text{diff} = I^+ - I^- = I_{||}^\text{direct}\]
leaving an image containing only directly reflected light. Images of two sample scenes can be found in Fig.~\ref{fig:polcompare}. Column (a) shows the difference image $I_\text{diff}$. Columns (b) and (c) show the parallel and perpendicular components $I_{||}^\text{meas}$ (b) and $I_\perp^\text{meas}$ (c), respectively, where the latter visibly contains only indirectly scattered light.

\subsection{Bipolar color matching functions} \label{sec:color}
\begin{figure}[t]
\begin{center}\includegraphics[width = 0.48\columnwidth]{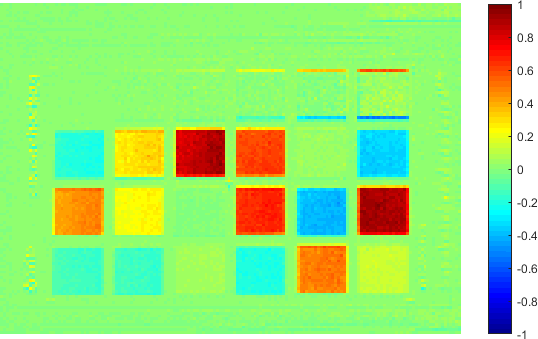}\hfill\includegraphics[width = 0.48\columnwidth]{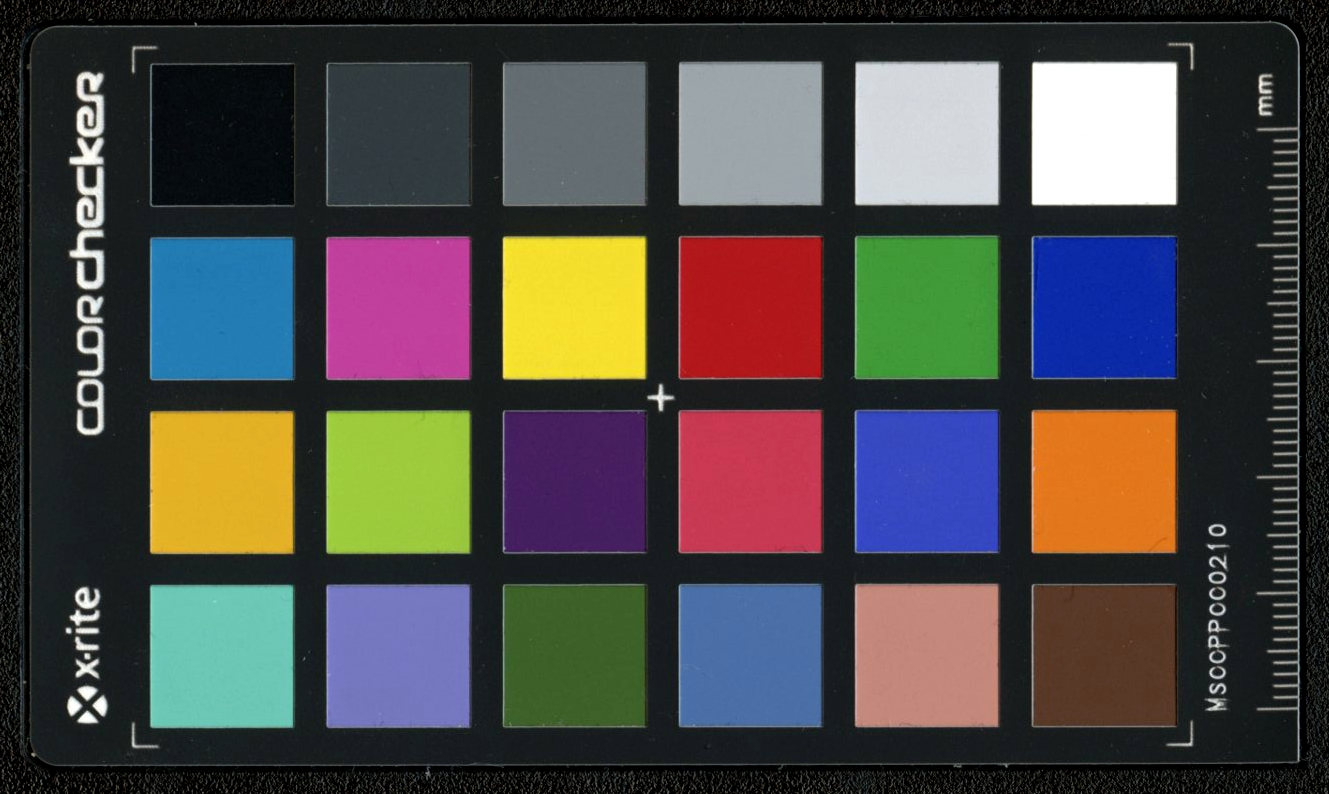}\vspace{-1mm}%
\caption{Left: Difference image of a color calibration chart, taken using PMD setup using alternating illumination of red and blue light. Reddish and bluish color patches obtain values on the opposite ends of the scale. Patches whose reflection spectra do not favor either red or blue obtain a value of approximately zero. Edges of some patches appear exaggerated due to partial shadowing of the light sources. Right: RGB scan taken of the color chart.}
\vspace{-1mm}
\label{fig:colorchecker}
\end{center}
\end{figure}

Being able to characterize and classify materials is important in many applications. Liu and Gu~\shortcite{liu2014discriminative} proposed to use discriminative illumination, or optimized pairs of spectro-angular illumination patterns to classify materials on a per-pixel basis. Here, we adopt the spectral aspect of this work, using our PMD setup to construct an active camera that discriminates between objects of red and blue reflectance in a single shot. By equipping L1 with red and L2 with blue LEDs, we obtain a \emph{bipolar} color camera that measures a positive response for objects that are predominantly red, and a negative response for bluish objects. Fig.~\ref{fig:colorchecker} shows an example measurement taken of the X-Rite ColorChecker, where the positive or negative response in the colored patches can clearly be seen. Patches that reflect red and blue to equal parts, like the greyscale, result in a response that is approximately zero. 

This example demonstrates the applicability of snapshot difference imaging to discriminative color imaging (e.g., red--blue). We envision this capability to facilitate interesting and novel approaches to image segmentation and classification or to enable direct sensing of primary colors with bipolar matching functions, like the red primary $\bar r(\lambda)$ in the CIE 1931 RGB color space\footnote{\url{https://commons.wikimedia.org/wiki/File:CIE1931_RGBCMF.svg}}.

\subsection{Depth edge and directional gradient imaging}

\begin{figure}[tbp]
\centering
\hspace*{-0.5mm}\begin{overpic}[width = 0.4\columnwidth]{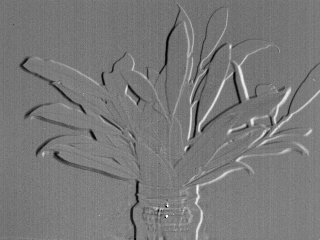}
\put (2, 2) {\textcolor{white}{(a)}}
\end{overpic}\hspace{1mm}
\begin{overpic}[width = 0.4\columnwidth]{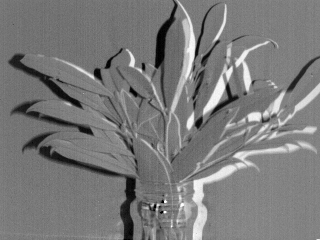}
\put (2, 2) {\textcolor{white}{(b)}}
\end{overpic}\\\vspace{1.5mm}
\begin{overpic}[width = 0.4\columnwidth]{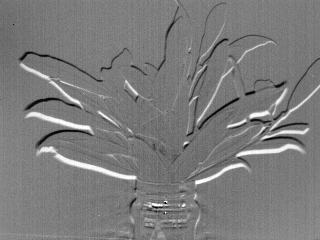}
\put (2, 2) {\textcolor{white}{(c)}}
\end{overpic}\hspace{1mm}
\begin{overpic}[width = 0.4\columnwidth]{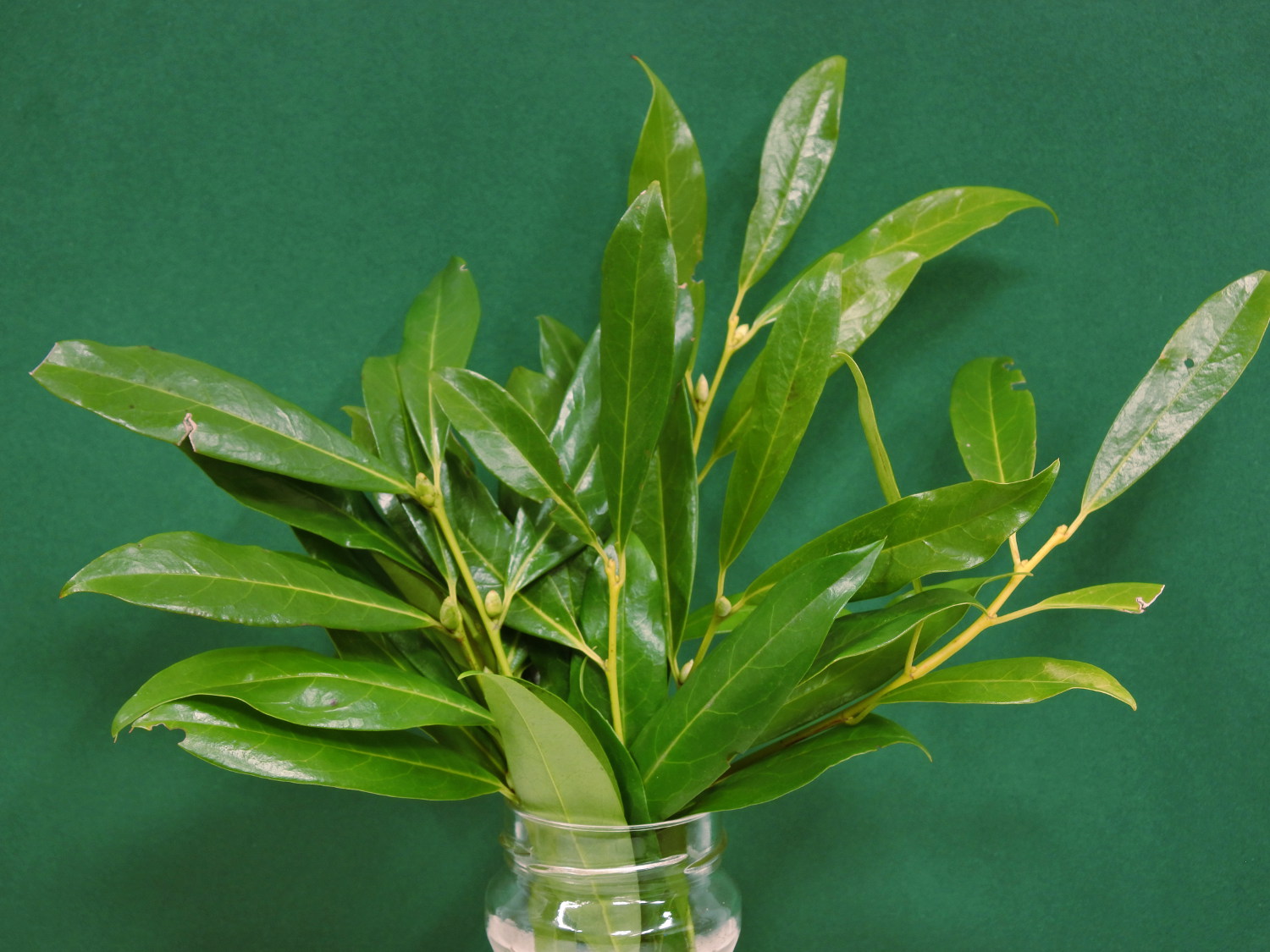}
\put (2, 2) {\textcolor{white}{(d)}}
\end{overpic}
\vspace{-1mm}
\caption{Horizontal (a)/(b) and vertical (c) alignment of the light sources creates corresponding gradient images. The distance between the light sources determines the width of the edges in the image: short distance (a) vs. larger distance (b).}
\label{fig:dir_plants}
\end{figure}

\begin{figure}[tbp]
\centering
\includegraphics[width = 0.45\columnwidth]{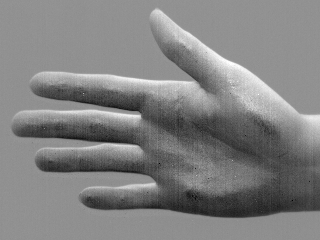}
\vspace{-1mm}
\caption{Directional difference image acquired with two light sources in wide vertical spacing.}
\label{fig:dir_hand}
\end{figure}
It is often hard to deduce the structure and shape of three-dimensional objects from conventional photographs, as they may show low contrast between spatially distinct features of the object. Illuminating the object from two different angles, however, can unveil the depth structure of a scene and facilitate, for instance, a segmentation of the image. Similarly to Raskar et al.~\shortcite{Raskar:2004}, our setup can be used to produce directional gradient images of a scene, visualizing depth continuities as shown in Fig.~\ref{fig:dir_plants}. In this mode of operation, two identical light sources of opposite polarity are placed on opposite sides of the sensor. Whenever a depth discontinuity shadows one of the light sources, the resulting image displays positive or negative values. All other pixels obtain a value around zero. By varying the distance between the light sources, different edge widths are obtained. As the light source separation approaches the distance between scene and camera, the system records shading images like Fig.\ref{fig:dir_hand}. Similarly to Woodham's photometric stereo method \shortcite{woodham1980photometric}, they could be used to estimate the surface orientation of an object.

\paragraph{Comparison of single- and two-shot edge imaging}
One of the key advantages of snapshot difference imaging is that it is immune to scene motion, whereas multi-shot techniques typically suffer from alignment issues when objects are rapidly moving. To illustrate this, we recorded two image sequences of a moving scene (bonsai tree shaken by wind) at the same frame rate of 60 frames per second. In Sequence 1, we used snapshot difference imaging with both light sources active; for Sequence 2, we alternated between LS1 and LS2, and digitally computed difference images between successive frames. As the results (provided as supplemental video) show, the single-shot difference images are significantly clearer with more consistent leaf shapes than the two-shot ones, and virtually free of ghosting artifacts. On the other hand, the single-shot images show a slight increase in fixed-pattern noise.


\subsection{Spatio-temporal gradient imaging}
A feature of difference imaging is its capability to extract essential information from heavy streams of image data. Here, we use our setup to implement cameras that selectively sense spatial or temporal changes in the input, opening use cases such as machine vision and data compression \cite{lichtsteiner2008temporalcontrast}. 

\paragraph{Spatial gradient.} 
\begin{figure}[tbp]
\centering
\includegraphics[width=0.75\columnwidth]{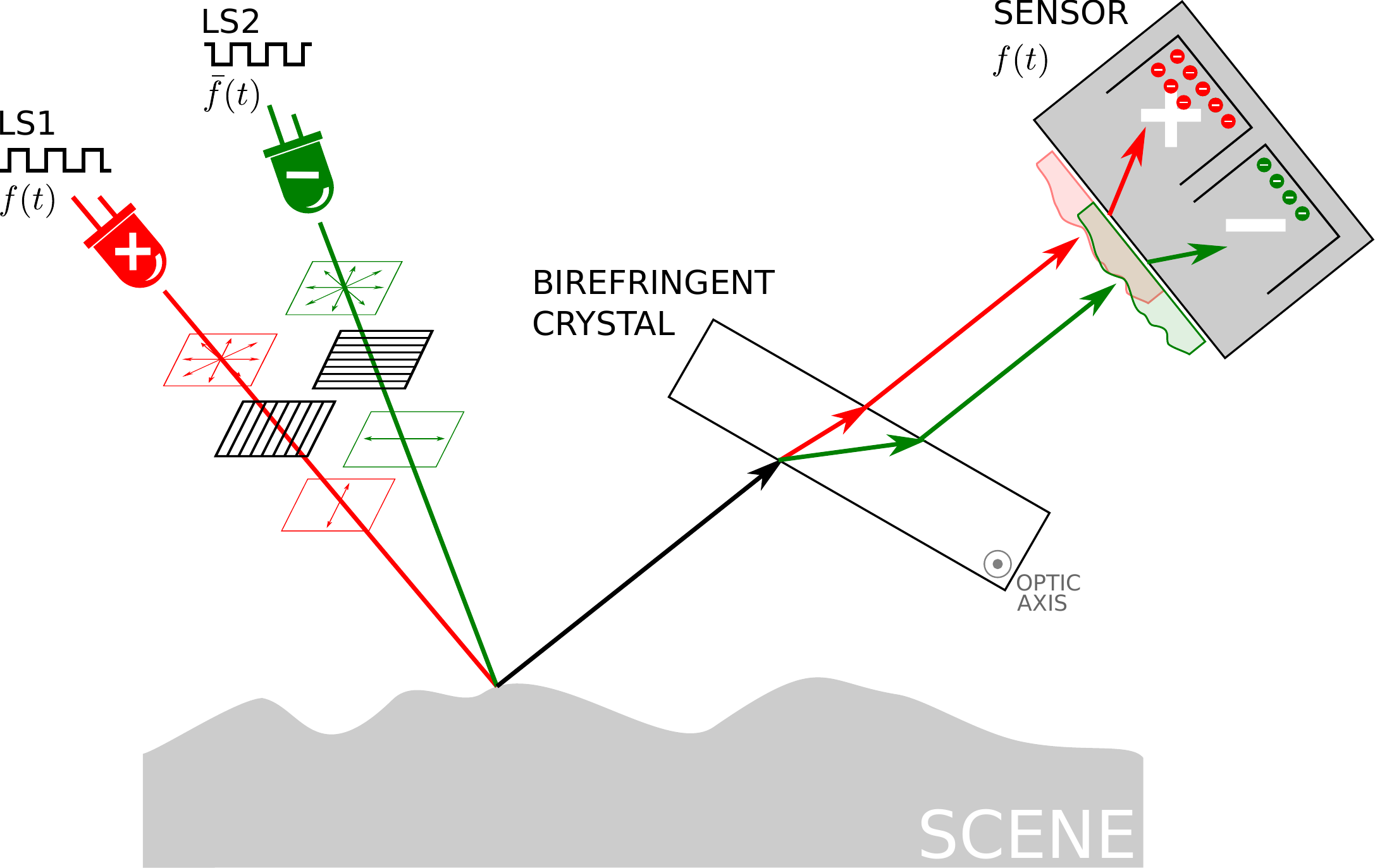}%
\vspace{-1mm}
\caption{A birefringent crystal is placed between the sensor and the scene; the scene is illuminated using identical light sources that are polarized in perpendicular angles. One of them is operated in phase (+) and the other in opposite phase (-) with the sensor. In direct reflections which preserve the polarisation, light from the light sources will be refracted in different angles inside the birefringent crystal and hence undergo a relative shift.}%
\label{fig:crystal_sketch}%
\end{figure}
\begin{figure}[tbp]
\centering
\includegraphics[width = 0.5\columnwidth]{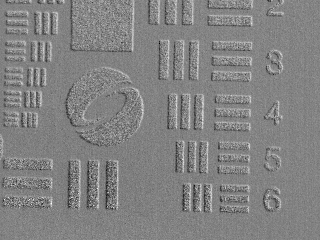}
\vspace{-1mm}
\caption{TI camera spatial gradient image of an aluminum resolution chart. In horizontal direction, the markers show black and white edges. The markers reflect more light than the surrounding area, which is why the  noise in the difference image is higher inside the markers than outside.}
\label{fig:crystal_edges}
\end{figure}
We devise an optical setup that, in combination with a snapshot difference imager, performs edge detection in analog hardware. The key is to introduce a small spatial displacement between the images $I^+$ and $I^-$, so the net image becomes the difference between two shifted copies of the scene. While this could in principle also be done through a mechanical element in the optical path (similar to active image stabilization in photography), we only add optically passive components to our setup.
In particular, we use oppositely polarized light sources as in Section~\ref{sec:direct_global}. Instead of the analyzing filter on the lens, we place a birefringent crystal immediately on top of the sensor, behind the camera lens. We determined that an undoped YVO4 crystal, 1\,mm thick and inclined by 20$^\circ$ with respect to the optical axis, causes a displacement between light of different polarization directions by about 15\,\textmu m, or one pixel of the TI sensor. For a polarization-preserving scene, this setup produces two identical images on the sensor area, displaced by one pixel and with opposite polarity. Uniform areas in the image cancel out in this difference image, while edges are detected as non-zero response (positive or negative depending on the direction). Figure~\ref{fig:crystal_edges} shows a gradient image of a planar aluminum resolution chart, recorded in a single shot using the TI setup. 
\begin{figure}[tbp]
\centering
\includegraphics[width = 0.7\columnwidth]{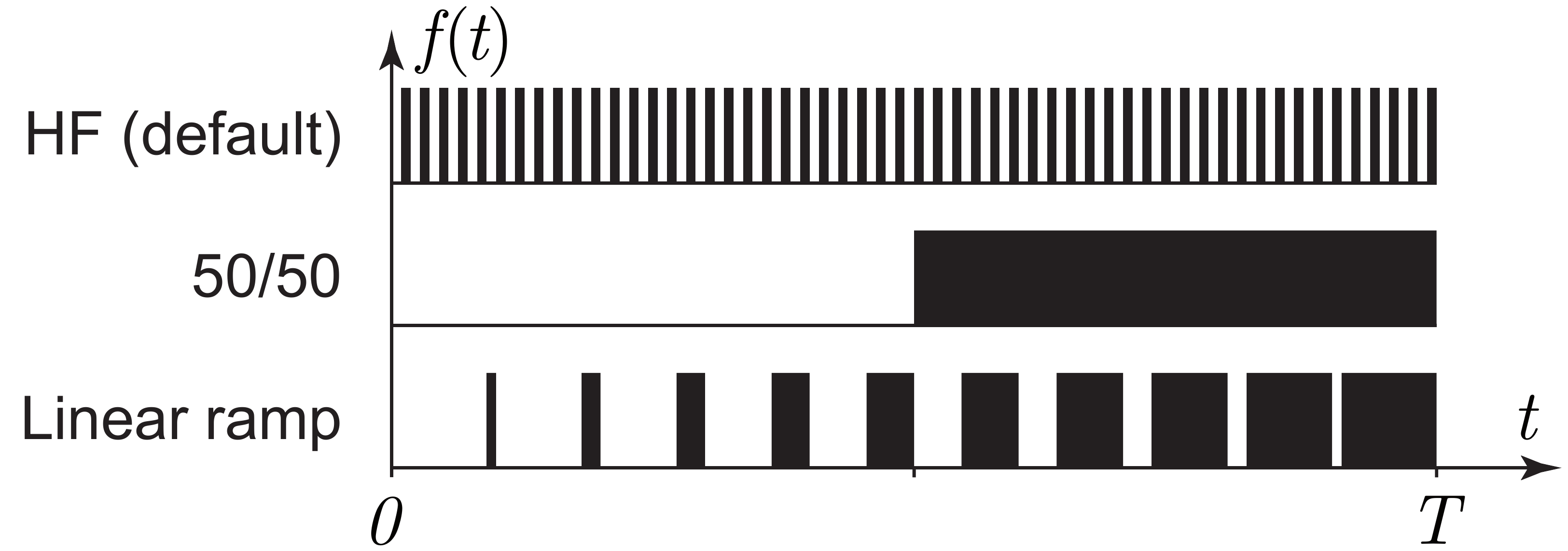}
\vspace{-3mm}
\caption{Example filters (modulation patterns) for use with our system. By default, the sensor is modulated at high frequency (top row). For the analog computation of temporal gradients, we use the 50/50 pattern with only one transition per exposure interval. This is implemented by modulating the sensor with a delayed version of the camera's ENABLE signal (cf.~Fig.~\ref{fig:setup}).}
\label{fig:tempgradpatterns}
\end{figure}
\begin{figure}[tbp]
\centering
\includegraphics[width = 0.38\columnwidth]{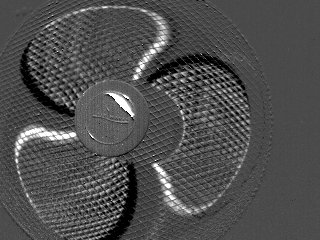}\hspace{4mm}
\includegraphics[width = 0.38\columnwidth]{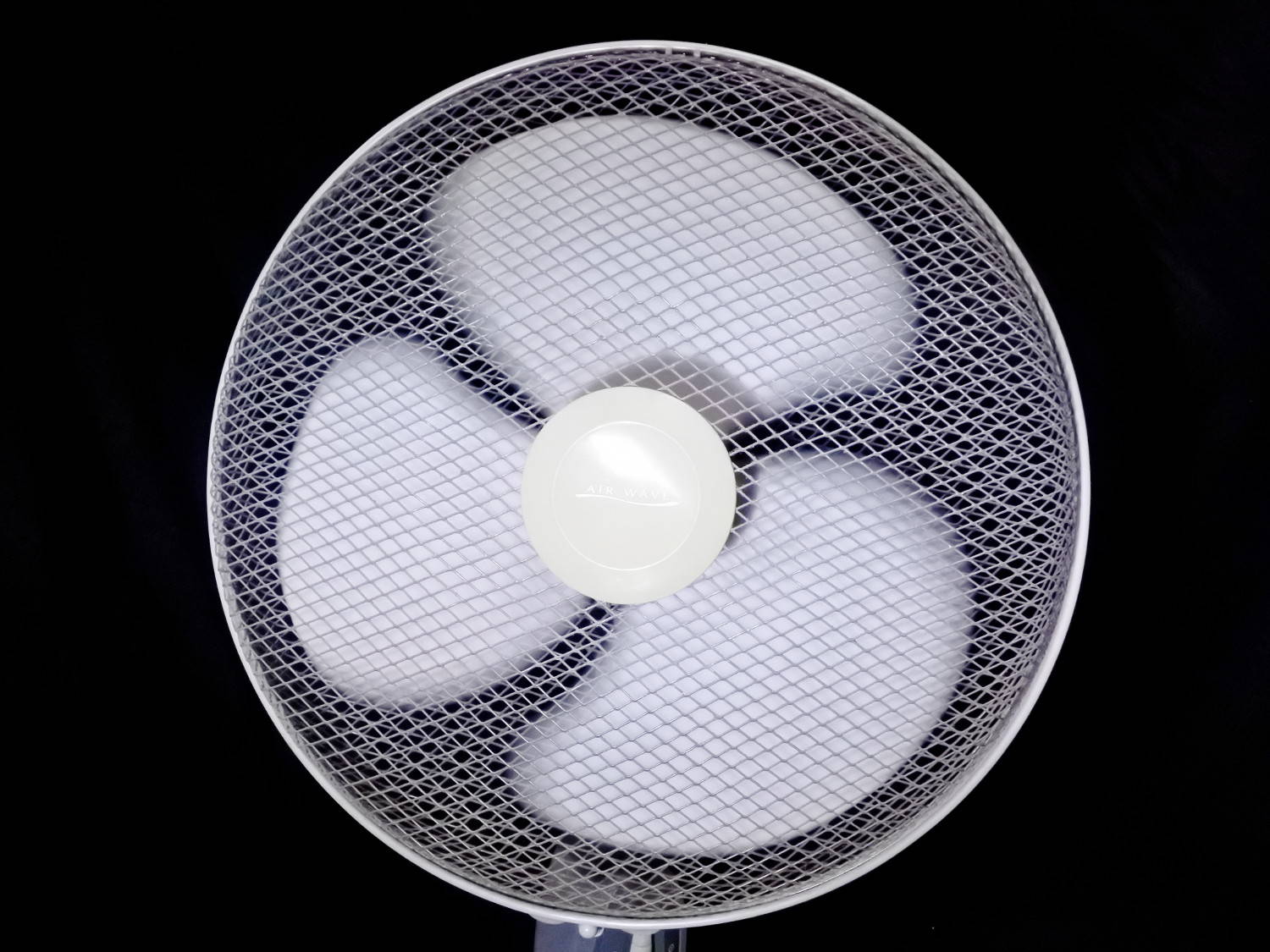}
\vspace{-1mm}
\caption{TI camera temporal gradient image (left) of a rotating fan (RGB image on the right). From the color gradient of blade edges the rotation direction of the fan is identifiable as clockwise.}
\label{fig:tempgrad_fan}
\end{figure}
\paragraph{Temporal gradient.} We conclude with an example for our difference imaging approach that can even be used without active illumination. 
So far, we modulated the sensor with a high-frequent square wave at 50\% duty cycle, which effectively made the sensor insensitive to ambient light. We now introduce a bias by choosing an asymmetric modulation pattern (Figure~\ref{fig:tempgradpatterns}). Light that arrives at the beginning of the exposure will now contribute to $I^-$, and light that arrives near the end will contribute more to $I^+$. In doing so, we make the camera sense temporal changes of intensity: pixels that receive more light during the second half of the exposure than during the first half appear as positive pixel values and vice versa. Figure~\ref{fig:tempgrad_fan} shows an image thus captured of a rotating fan. From the temporal gradient image, the direction of rotation can be identified by the black and white edges of the blades. 

Another example is shown in Fig.~\ref{fig:teaser}(b), where white pellets are shown falling on the ground. The direction of motion is visible in the temporal gradient image: Pellets falling to the ground feature positive values (red) on the bottom end and negative  values (blue) on the top end. Those that have bounced off the ground and fly back up (as seen in the right part of the image) have reversed shading. Pellets lying still on the ground are barely visible. 

The exposure time used for this method is 1\,ms, thus transferring this method to a conventional camera would correspond to a required frame rate of 2000\,fps.

\subsection{Quantitative Noise Analysis}

\begin{figure}[tbp]
\centering
\includegraphics[width=0.8\columnwidth]{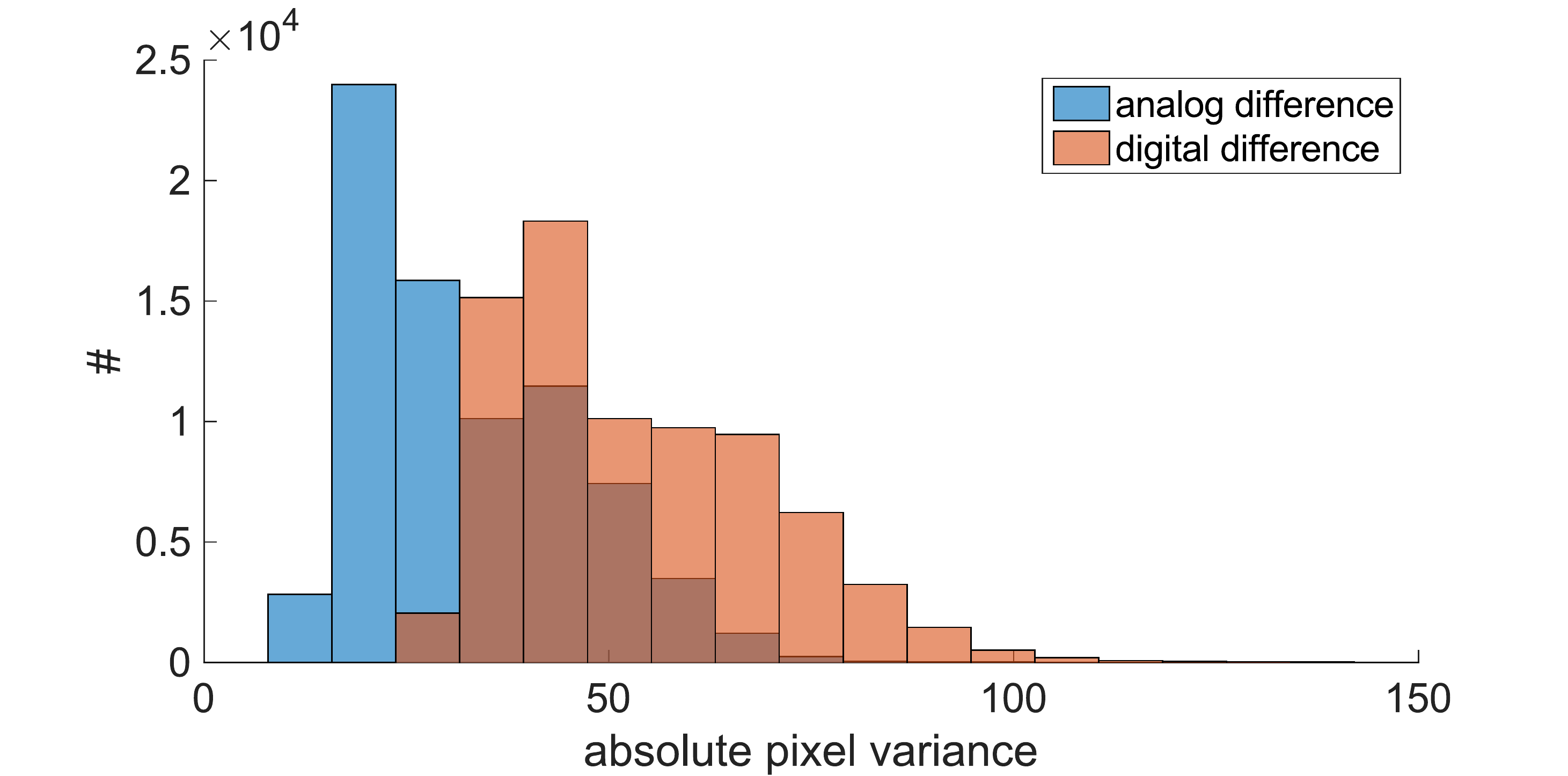}\vspace{-1mm}%
\caption{Histograms of the absolute pixel variance in the analog and digital difference images of the scene depicted in the bottom row of Fig.~\ref{fig:polcompare}.}%
\vspace{-1mm}
\label{fig:histo_variance}%
\end{figure}

In contrast to conventional cameras, our snapshot difference imaging approach performs only one read-out operation in the process of obtaining a difference image, since the differencing operation is performed before the readout. Hence, assuming shot noise and read noise as the main contributions to the measurement uncertainty, a pre- and post-ADC difference image are expected to suffer from different noise levels: 
\begin{eqnarray}
\label{eq:pre_post_noise}
\begin{aligned}
&\left(\sigma_\text{diff}^\text{post}\right)^2 = \eta^2\sigma_+^2 + \sigma_\text{read}^2 + \eta^2\sigma_-^2 + \sigma_\text{read}^2 ,\\
&\left(\sigma_\text{diff}^\text{pre}\right)^2 = \eta^2\sigma_+^2 + \eta^2\sigma_-^2 + \sigma_\text{read}^2.
\end{aligned}
\end{eqnarray}
To compare the relative performance of snapshot difference imaging with two-shot, post-capture difference imaging under otherwise identical conditions, we acquired three image sequences of a still scene, each $N$ frames long: one sequence with both light sources activated (also shown in Fig.~\ref{fig:polcompare}(a)), and two more with only LS1 or LS2 turned on, respectively (Fig.~\ref{fig:polcompare}(b) and Fig.~\ref{fig:polcompare}(c)). We then used the data acquired with separate light sources to compute another set of difference frames.
As a measure for the signal quality of the difference frames, we computed for each pixel the variance across the $N$ recorded frames, and plotted the values for all pixels in a histogram. As Fig.~\ref{fig:histo_variance} illustrates for a case with $N = 100$ frames, the noise in the snapshot difference image is significantly lower than in the post-capture difference image. 

\begin{figure}[tbp]
\centering
\includegraphics[width=0.49\columnwidth]{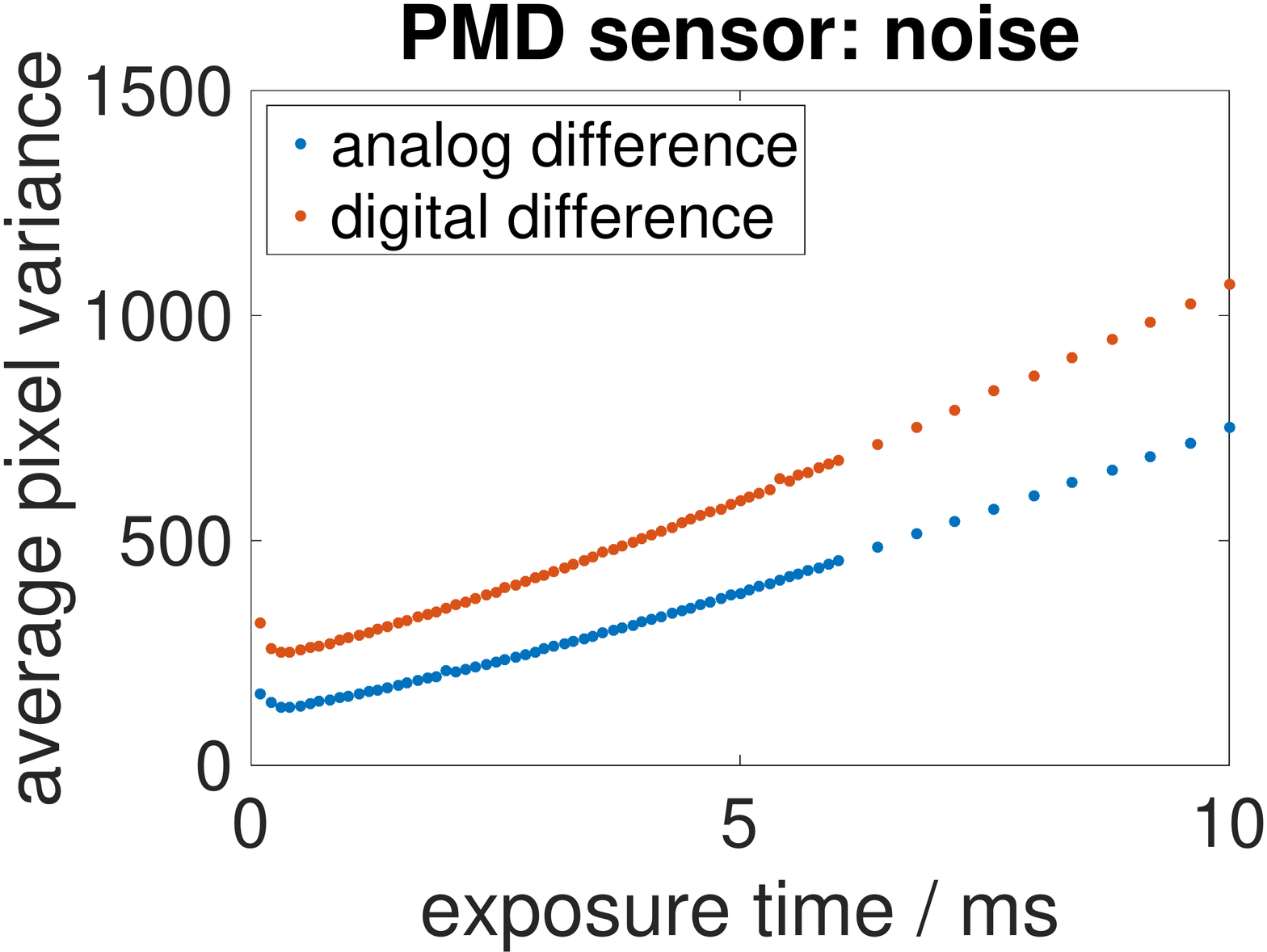}
\includegraphics[width=0.49\columnwidth]{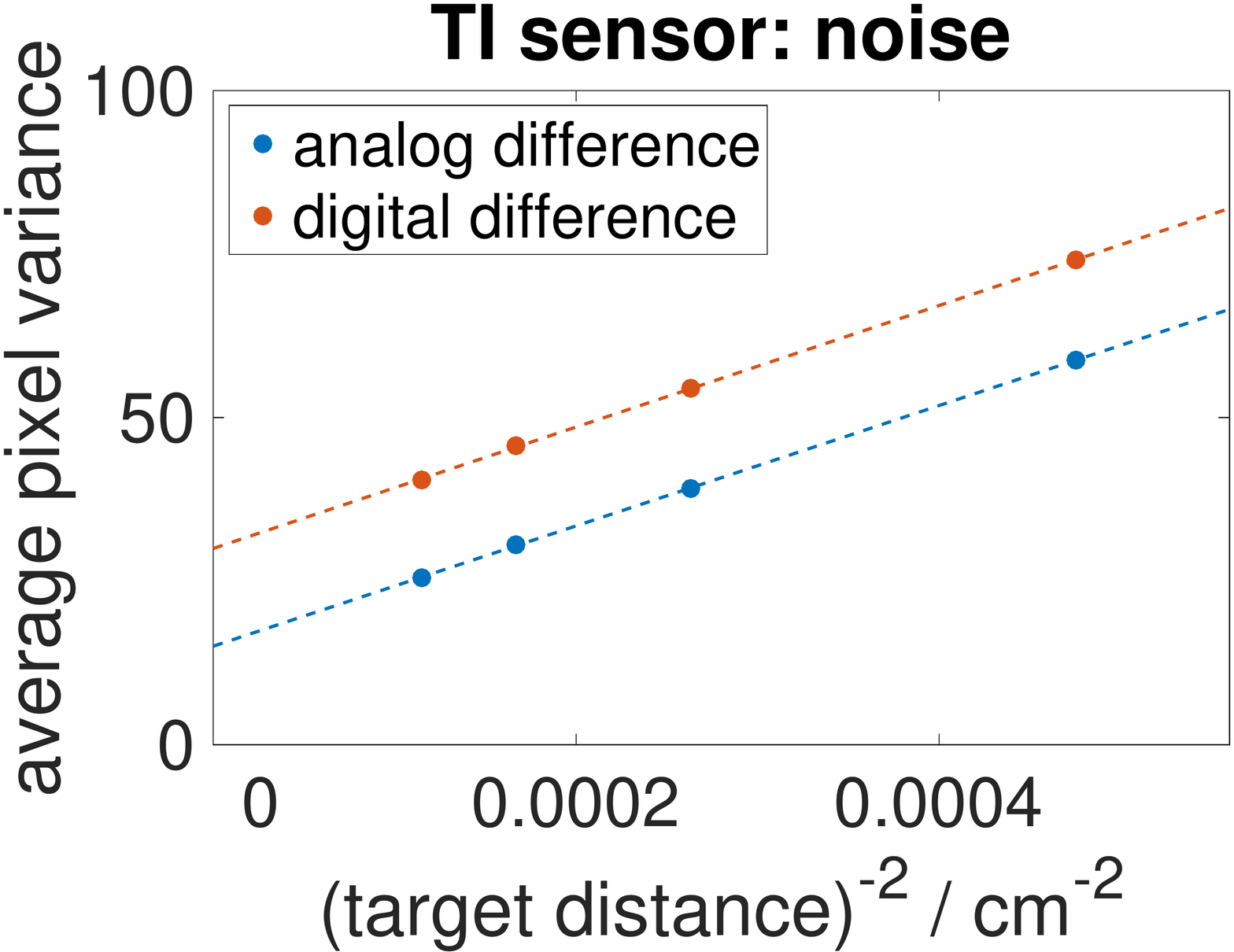} \\
(a)\hspace{0.45\columnwidth}(b)\vspace{-3mm}
\caption{Variance of each pixel, averaged over the whole difference image, for the PMD sensor (a) and the TI sensor (b). The values were obtained from 400 frames each. The lines in (b) are linear fits to the data.}%
\label{fig:exp_variance}%
\end{figure}

Fig.~\ref{fig:exp_variance} shows the pixel variance (averaged over all image pixels) in dependence of the intensity of the incident light. For the PMD camera, we varied the exposure time of otherwise identical shots using red and blue illumination. Since the TI sensor does not allow adjustment of the exposure time, we placed a white, homogeneous target in different distances from the camera and light sources in order to obtain different intensities. In both cases, the post-ADC difference image consistently shows higher noise. For low light intensities (short exposure times or large target distances, respectively), the shot noise of both pixel buckets tends to zero, so in Eq.~\ref{eq:pre_post_noise}, only the read noise terms remain and one has $$\left(\sigma_\text{diff}^\text{post}\right)^2 = 2 \left(\sigma_\text{diff}^\text{pre}\right)^2.$$ The data depicted in Fig.~\ref{fig:exp_variance} supports this expectation as the lowest measured variance values for the PMD sensor are $$\left(\sigma_\text{diff}^\text{post}\right)^2_\text{min} = 251.37,\quad \left(\sigma_\text{diff}^\text{pre}\right)^2_\text{min} = 127.90$$ (ratio 1.97) and the extrapolated (via a linear fit) lowest variance values for the TI sensor are $$\left(\sigma_\text{diff}^\text{post}\right)^2_\text{min} = 29.99 \pm 0.30,\quad \left(\sigma_\text{diff}^\text{pre}\right)^2_\text{min} = 15.05 \pm 0.22$$ $(\text{ratio } 1.99 \pm 0.37)$, which is in good agreement with a ratio of 2.\footnote{Since it is not possible to read out the pixel buckets separately, we cannot exclude the possibility that what we model as $\sigma^2_\text{read}$ is partly constituted of noise that is introduced by the process of taking the difference voltage of both pixel buckets. This would reduce the factor between post- and pre-ADC-differencing in a setup with an otherwise identical conventional sensor to a value between 1 and 2.}

\begin{figure}[tbp]
\centering
\begin{overpic}[height=31mm]{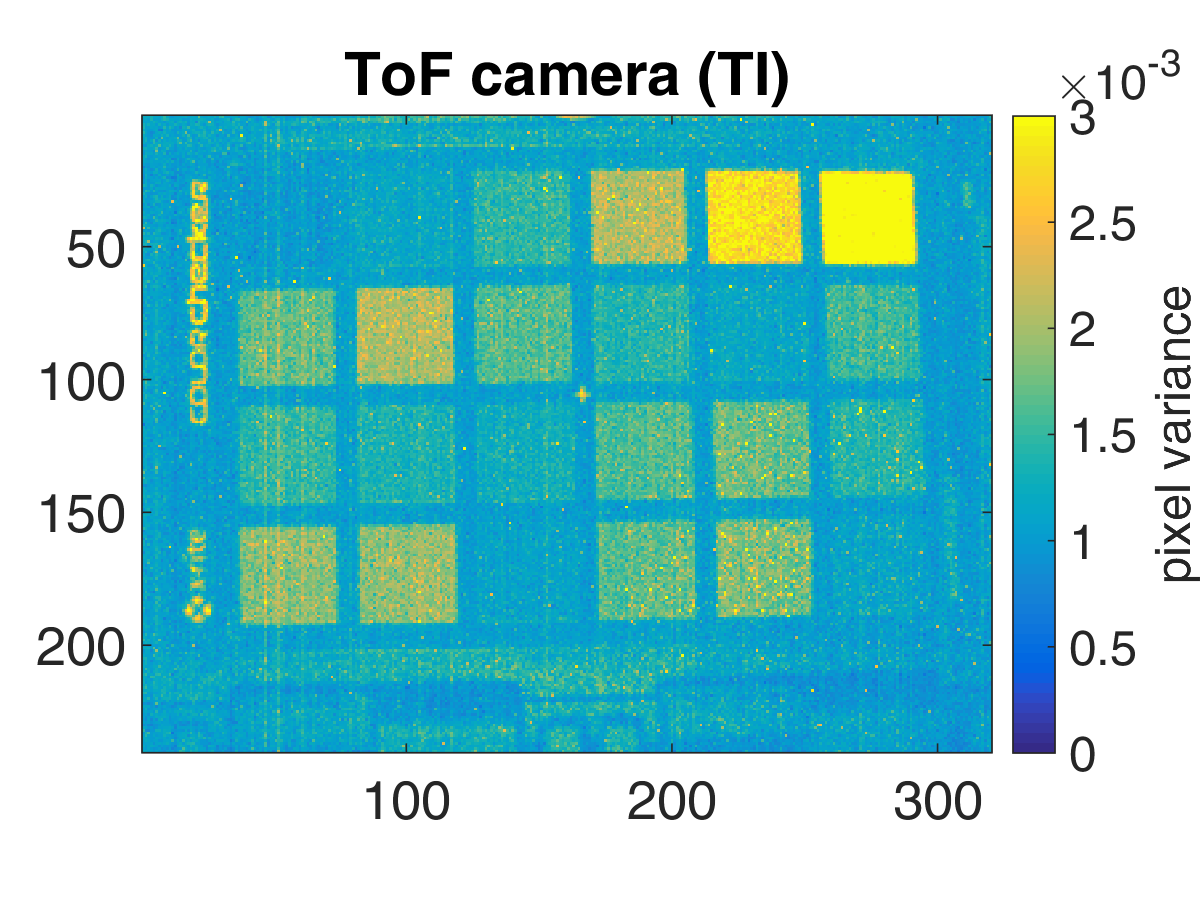}
\put (14, 15) {\textcolor{white}{(a)}}
\end{overpic}
\begin{overpic}[height=31mm]{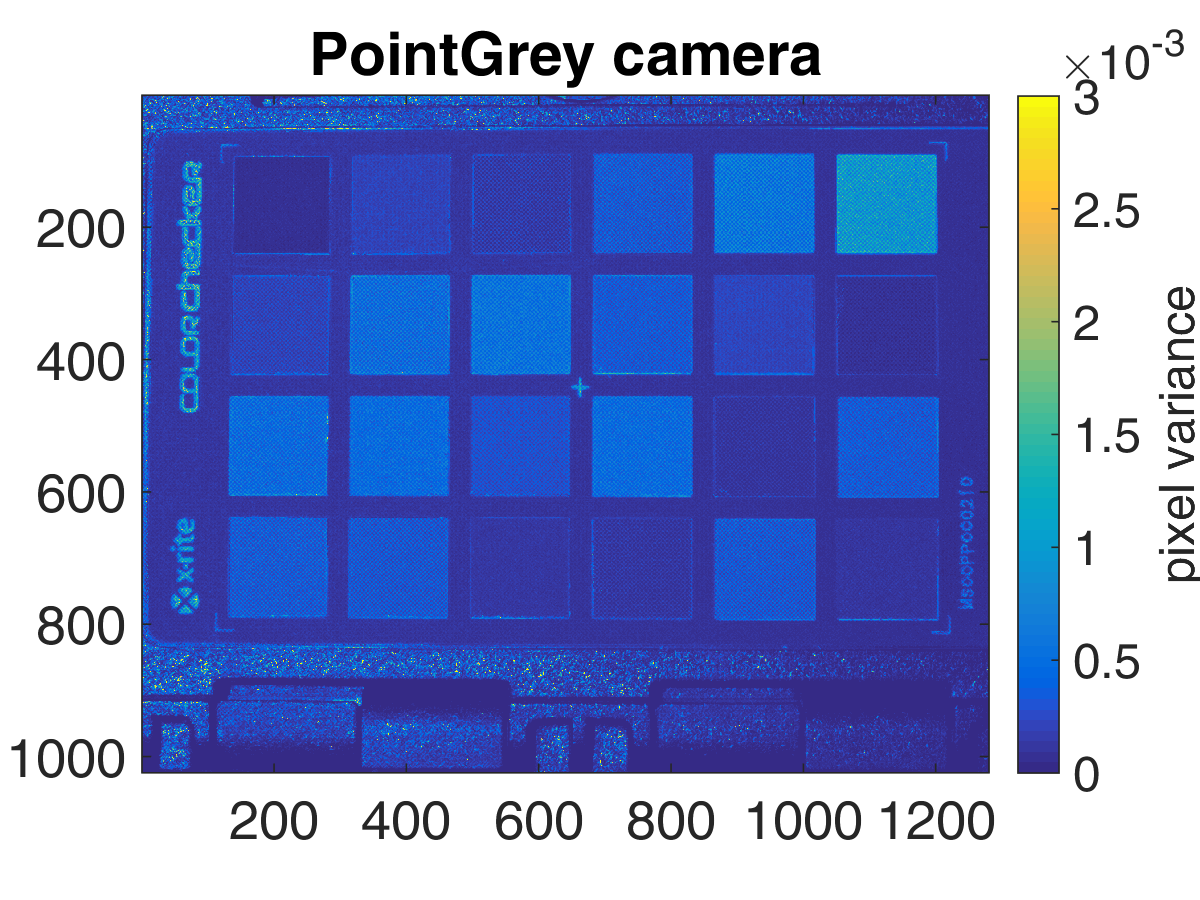}
\put (14, 13) {\textcolor{white}{(b)}}
\end{overpic}\\
\vspace{-4mm}%
\caption{Pixel variance in difference images of a color chart recorded with a TI-ToF-camera (a) and a PointGrey Flea3 camera (b). The variances were calculated from 400 and 137 frames, respectively.}%
\label{fig:PG_compare}%
\end{figure}

In order to embed our setup into the context of existing camera hardware, we compared the noise level of the TI-ToF-camera to a conventional (PointGrey Flea3) camera.  Figure~\ref{fig:PG_compare} shows the variance of each pixel in a series of difference images of a color chart taken with both cameras. For the PG camera, two sets of images have been recorded and subtracted digitally. In order to make this comparison as fair as possible, we used the same LED light sources and  a shutter time of 1\,ms for the TI camera and 0.5\,ms for the two separate images taken with the PG camera. Evidently, the ToF camera shows higher noise and lower resolution than the PointGrey camera which is expected due to the much longer development history of conventional image sensors compared to ToF sensors.

\begin{figure}[tbp]
\centering
\begin{overpic}[width=0.49\columnwidth]{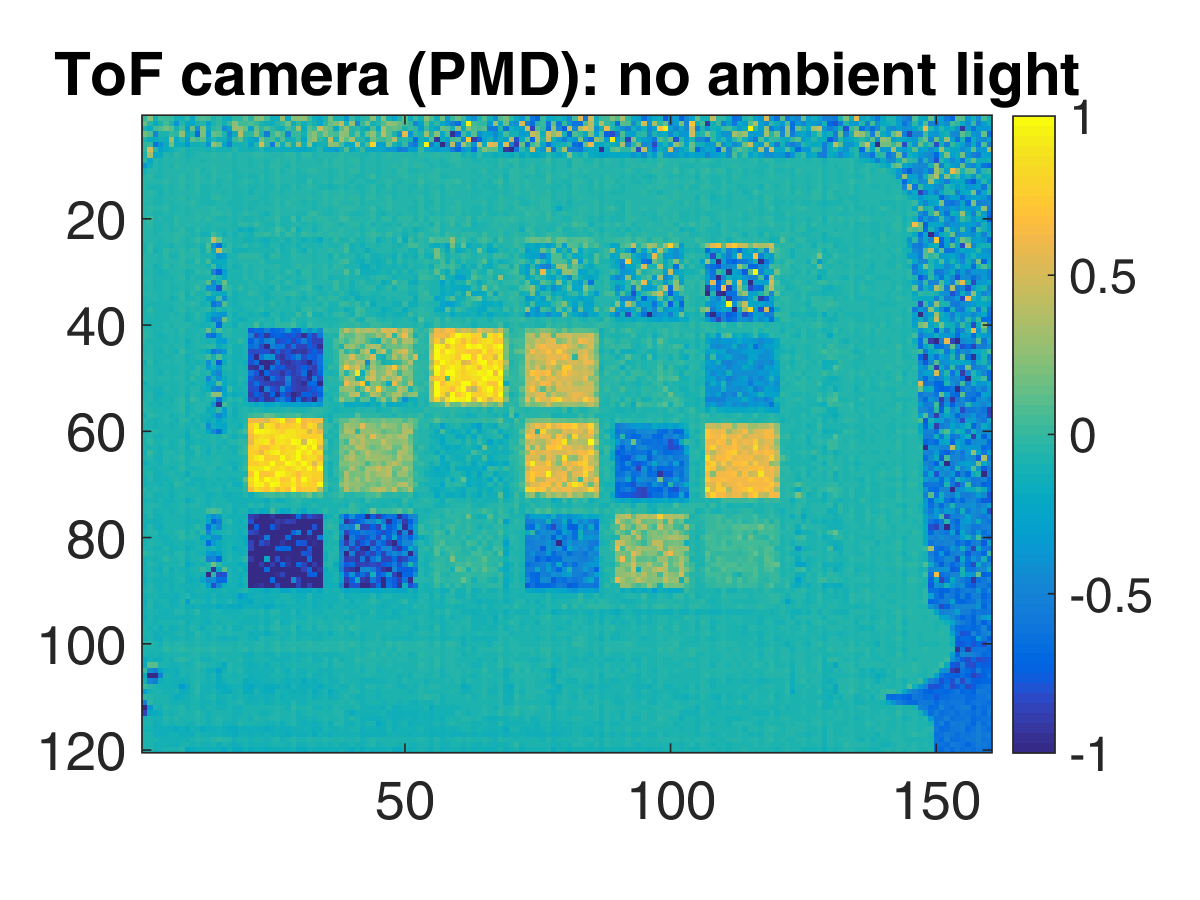}
\put (14, 15) {\textcolor{white}{(a)}}
\end{overpic}
\begin{overpic}[width=0.49\columnwidth]{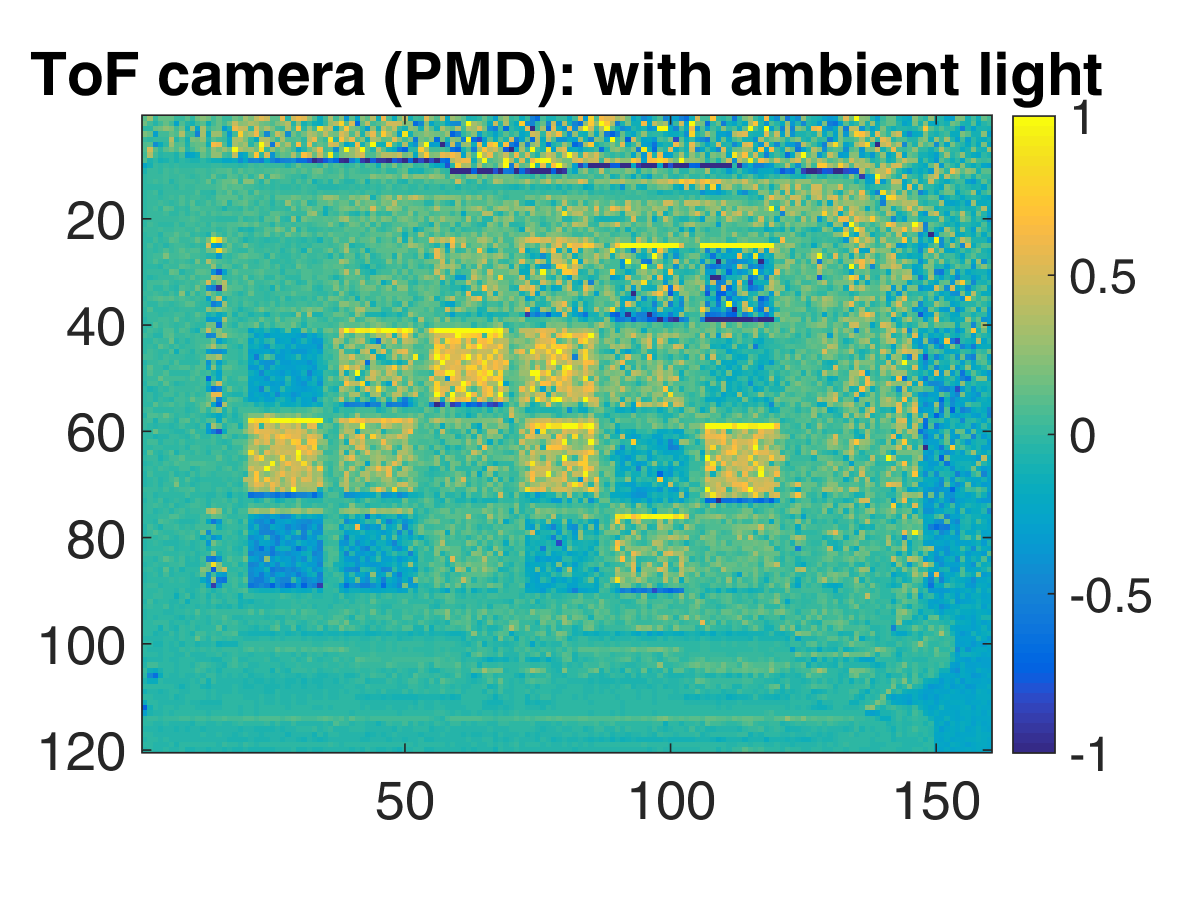}
\put (14, 15) {\textcolor{white}{(b)}}
\end{overpic} \\\vspace{-2mm}
\begin{overpic}[width=0.49\columnwidth]{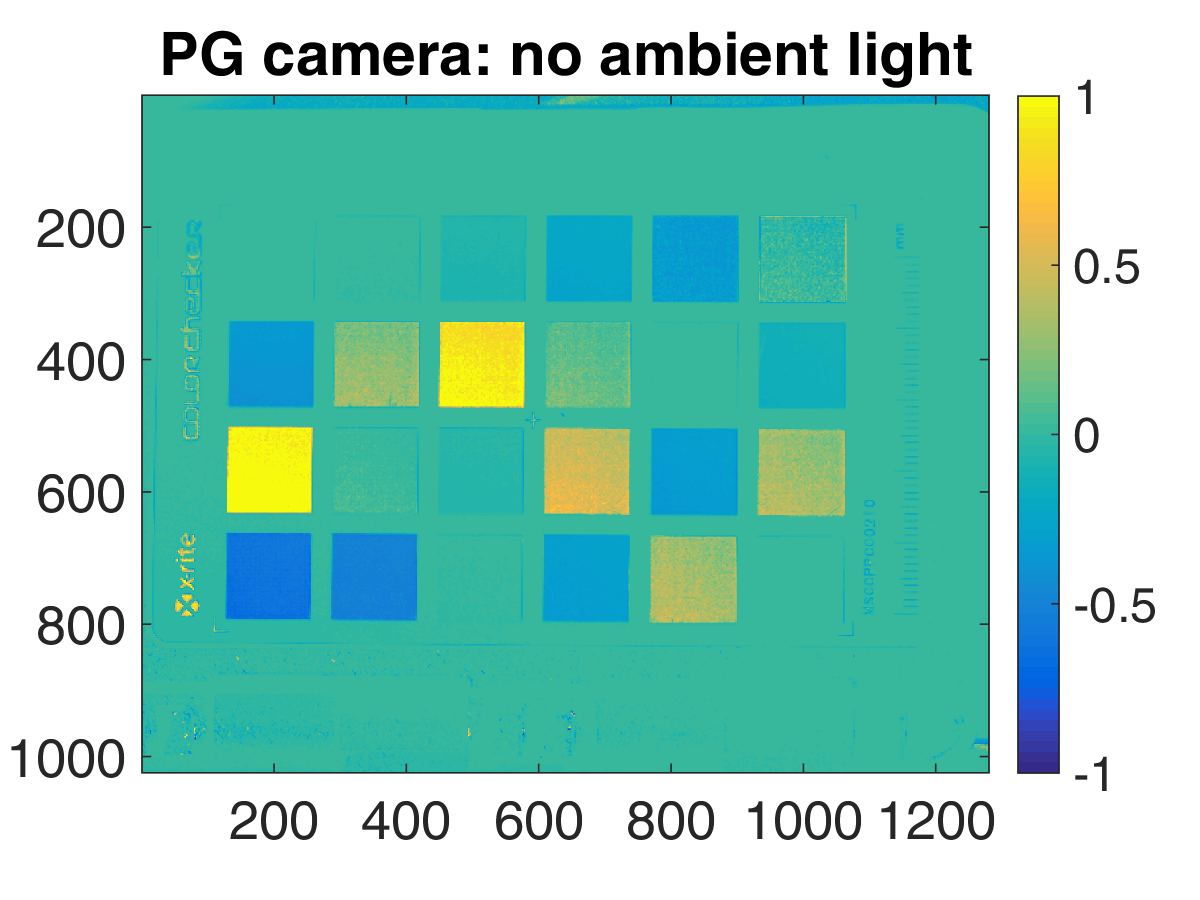}
\put (14, 13) {\textcolor{white}{(c)}}
\end{overpic}
\begin{overpic}[width=0.49\columnwidth]{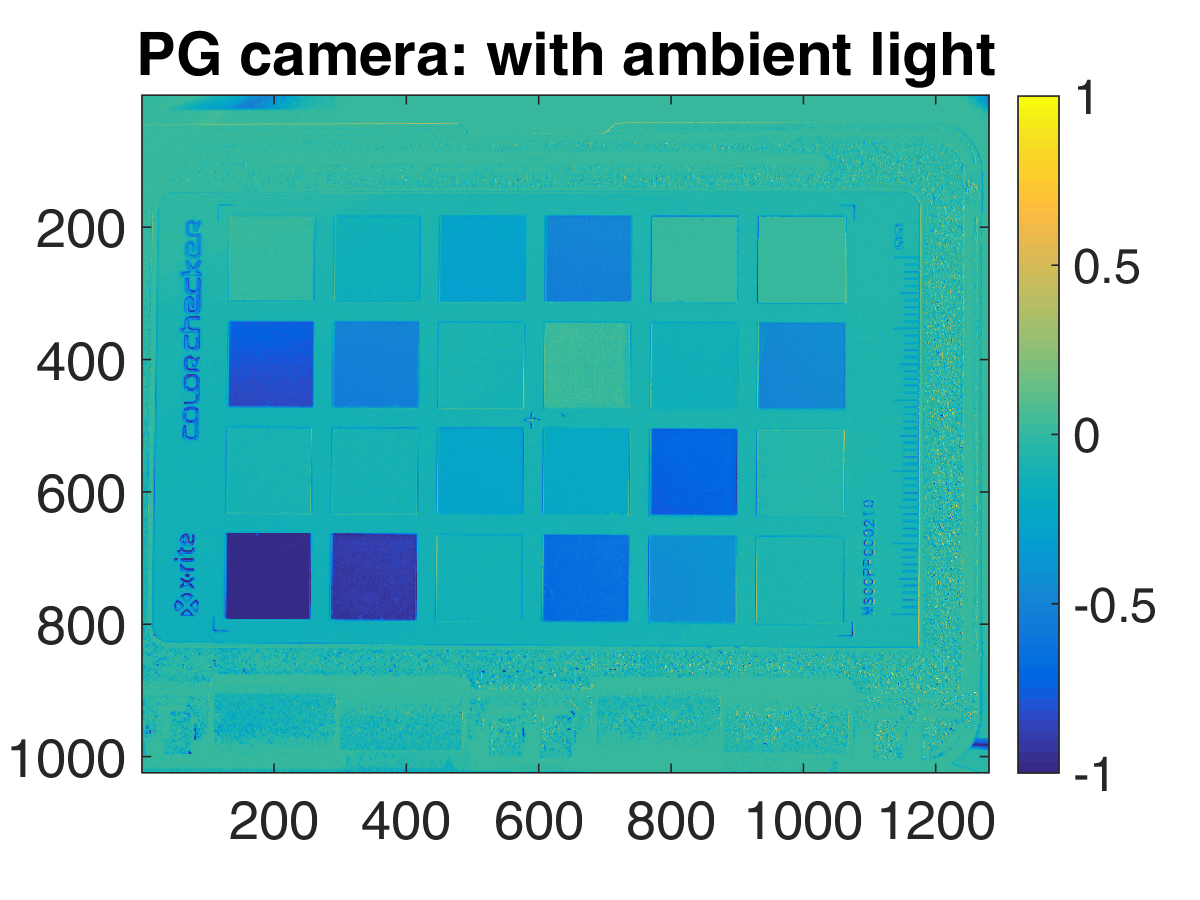}
\put (14, 13) {\textcolor{white}{(d)}}
\end{overpic} \\\vspace{-0.5mm}
\begin{overpic}[width=0.49\columnwidth]{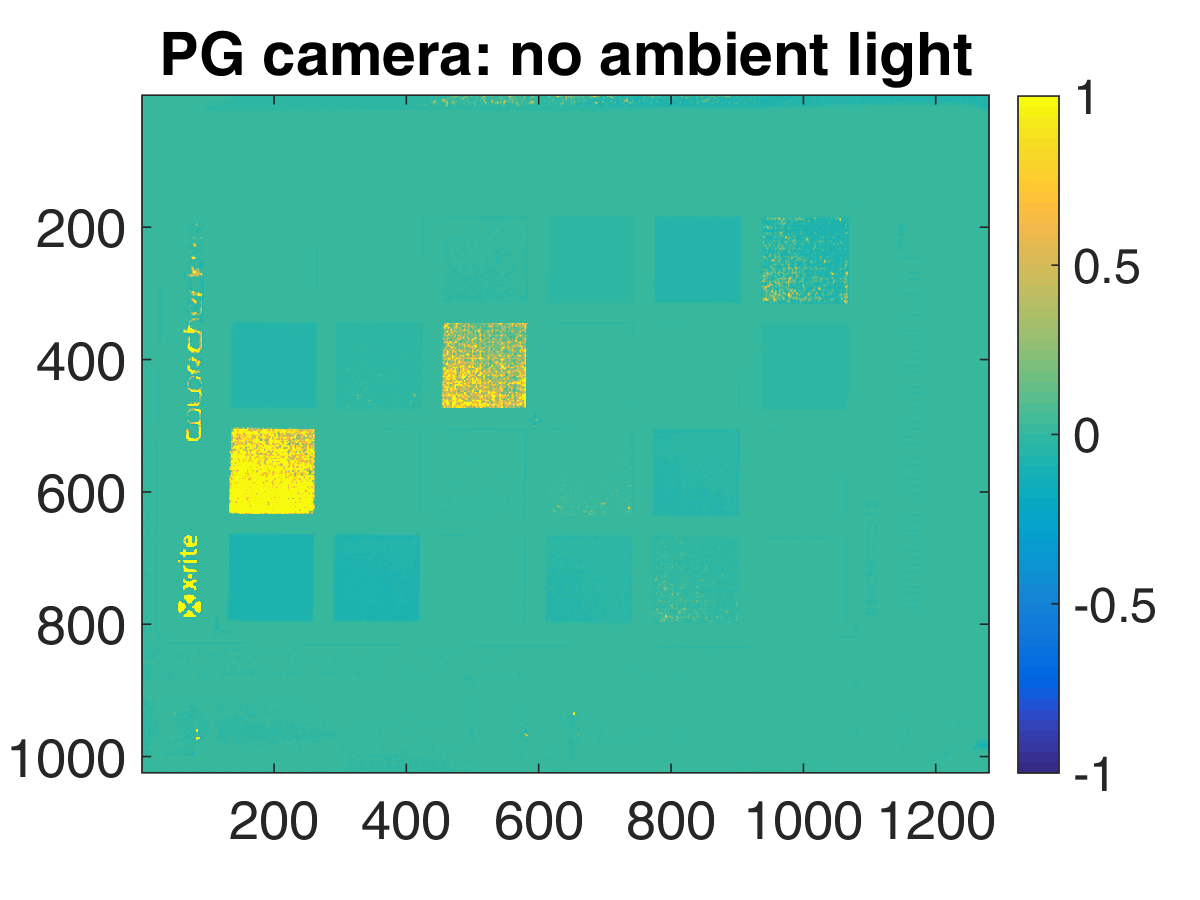}
\put (14, 13) {\textcolor{white}{(e)}}
\end{overpic}
\begin{overpic}[width=0.49\columnwidth]{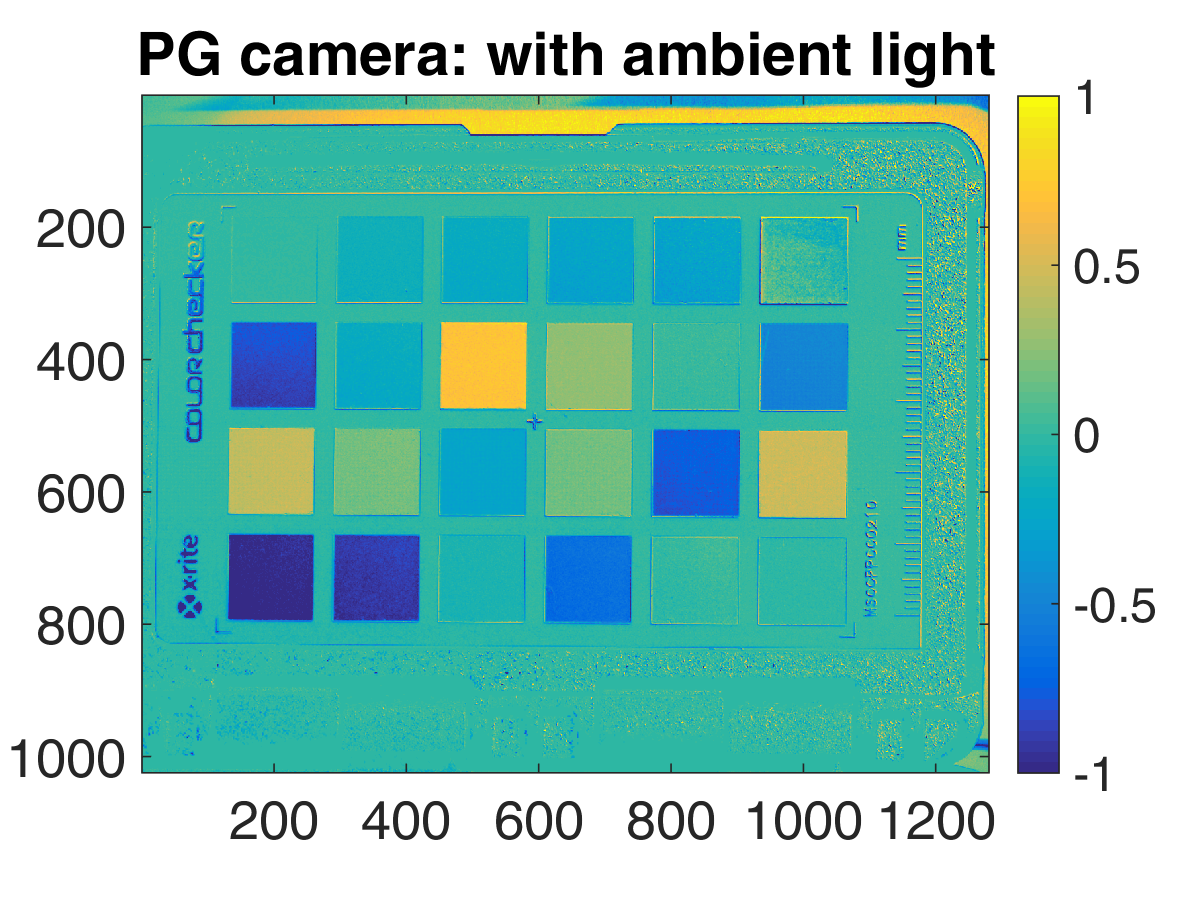}
\put (14, 13) {\textcolor{white}{(f)}}
\end{overpic}\\\vspace{-4mm}
\caption{Difference images without (left) and with (right) ambient illumination, acquired with a PMD ToF camera (top row) and a conventional PointGrey Flea3 camera with larger (center row) and smaller (bottom row) aperture setting.}%
\label{fig:SBI}%
\end{figure}

The special architecture of ToF sensors, however, allows to neglect photoelectric charges in the pixel buckets that have been produced by light that is not correlated with the sensor modulation, i.e. ambient light. Figure~\ref{fig:SBI} shows difference images of a color chart obtained with a PMD ToF sensor and a PG camera with and without ambient illumination of (several magnitudes) higher intensity than the active illumination. As shown in the upper row of Fig.~\ref{fig:SBI}, the additional light decreases contrast and increases noise in the difference image of the PMD sensor. With the PG camera, the sensor's dynamic range does not allow to find a setting that captures both the target setup with and without ambient light correctly. Thus, in subfigure~\ref{fig:SBI}d, several patches of the color chart obtain pixel values of zero, because two saturated pixels have been subtracted from each other. If the camera parameters are adjusted such that the image with ambient illumination is properly exposed, the image without ambient light is too dark to show meaningful values (bottom row of Fig.~\ref{fig:SBI}). While in both cases the image quality for the properly exposed images is clearly better with the PG camera, the PMD camera shows higher variability and adaptability in terms of ambient illumination. Since the ability to suppress ambient illumination is unique to the ToF sensor type, we suppose that it has the potential of increasing the range of possible applications for the snapshot difference imaging approach in contrast to conventional camera setups in the future.

\section{Recovering Two Images from a Single Difference Image}
\label{sec:separation}
In this section, we document an interesting side observation that falls directly out of the proposed image formation model for difference imaging. We can recover the two original images from a single difference image by exploiting the noise characteristics of both photon limited signals. According to Eq.~\ref{eq:skellammatrix}, the noise in each pixel of a difference image is dependent on the amount of charges stored in the individual wells, rather than the resulting difference value. Therefore, we can calculate the separate values $I^+$ and $I^-$ from the noise statistics (mean and variance of each pixel) of the difference image:
\vspace{-10pt}
\begin{equation}
\begin{pmatrix}
I^+ \\ I^-
\end{pmatrix}
=
H^{-1}
\begin{pmatrix}
\mu \\ \sigma^2
\end{pmatrix}.
\end{equation}
We propose three methods (M1--M3) to estimate these quantities:
\begin{enumerate}[leftmargin=*,label=M\arabic*:\!]%
	\item Analysis of a sequence of input image frames $I^{\vec x}_{1..N}$ taken under identical conditions:
	\begin{eqnarray*}
\textstyle	\mu^{\vec x}=\frac{1}{N}\sum_{i=1}^N I^{\vec x}_i\quad\text{and}\quad(\sigma^2)^{\vec x}=\frac{1}{N-1}\sum_{i=1}^N \bigl(I_i^{ {\vec{x}}} - \mu_i^{\vec x}\bigr)^2,
	\end{eqnarray*}
	where $I^{\vec x}_i$ denotes the pixel value at location $\vec x = (x,y)$ in the $i^\textrm{th}$ frame. 
	
	\item Patch-based analysis of a single pre-segmented image:
	\begin{eqnarray*}
\textstyle \mu^{\vec x}=\frac{1}{|P^{\vec x}|}\sum_{{\vec{x}}'\in P^{\vec x}}\!I^{ {\vec{x}}'}\quad \text{and}\quad(\sigma^2)^{\vec x}=\frac{1}{|P^{\vec x}|\!-\!1}\sum_{ {\vec{x}}'\in P^{\vec x}}\bigl(I^{ {\vec{x}}'} - \mu^{\vec x}\bigr)^2\!\!,
	\end{eqnarray*}
	where $P^{\vec x}$ denotes the set of pixels belonging to the same image segment (patch) as pixel $\vec x = (x,y)$.
	\item Analysis of a single image using a bilateral 
	filter:
	\begin{eqnarray*}
\textstyle	\mu^{\vec x}=\frac{1}{\sum_{\vec x'}w}\sum_{\vec x'}I^{\vec x'}w\quad\textrm{and}\quad
(\sigma^2)^{\vec x}=\frac{1}{\sum_{\vec x'}w}\sum_{\vec x'}\bigl(I^{\vec x'}-\mu^{\vec x}\bigr)^2w
	\end{eqnarray*}
 with the bilateral weight $w$ \cite{tomasi1998bilateral}
	\begin{equation*}
\textstyle	w(\vec x',\vec x,I^{\vec x'},I^{\vec x})=
e^{-{(I^{\vec x'}-I^{\vec x})^2}/{2\sigma^2_\textrm{range}}
}~e^{-{(\vec x'-\vec x)^2}/{2\sigma^2_\textrm{domain}}}.
	\end{equation*}
\end{enumerate}

\begin{figure}[tbp]
\hspace*{-0.7mm}\begin{overpic}[width = 0.36\columnwidth]{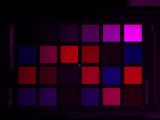}
\put(2, 3){\textcolor{white}{(a)}}
\end{overpic}\hspace{1mm}
\begin{overpic}[width = 0.36\columnwidth]{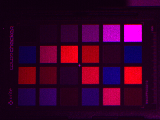}
\put(2, 3){\textcolor{white}{(b)}}
\end{overpic}\\\vspace{1mm}
\begin{overpic}[width = 0.36\columnwidth]{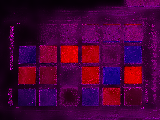}
\put(2, 3){\textcolor{white}{(c)}}
\end{overpic}\hspace{1mm}
\begin{overpic}[width = 0.36\columnwidth]{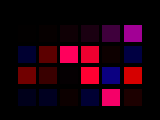}
\put(2, 3){\textcolor{white}{(d)}}
\end{overpic}
\vspace{-1mm}
\caption{Reconstruction of two color channels from a single exposure by exploiting photon statistics. (a) Ground-truth image combined from isolated measurements of red and blue illumination; (b) Reconstruction from 1000 difference images (Method 1); (c) Reconstruction from one single difference image (Method 3); (d) Reconstruction from one manually segmented image (Method 2).  All images were acquired with the PMD sensor.}
\label{fig:reco_rb}
\end{figure}

Figure~\ref{fig:reco_rb} shows the reconstructions of the individual blue and red channels from the difference image shown in Fig.~\ref{fig:colorchecker}a (for acquisition details, see section~\ref{sec:color}), obtained using M1--M3 without the read noise term ($\sigma^2_\textrm{read}\!:=\!0$). 100 dark frames were acquired, averaged and subtracted from the difference images before performing the reconstruction.  M1, here using $N\!=\!1000$ frames, delivers the best result. M2 and M3 sacrifice quality to separate the sources from one single difference image, which makes them suitable for fast moving target scenes. M2 yields the next-best reconstruction regarding color quality and particularly the gray scale (top row of patches), but it requires flat homogeneous image regions (pre-segmented by hand). M3 uses a bilateral filter to weight down dissimilar pixels when computing mean and variance. This reduces the overall estimated variance and introduces bias; nevertheless, this algorithm would be simple enough for real-time applications.

\begin{figure}
\centering\small
	\begin{tabularx}{0.88\linewidth}{cc}
	\includegraphics[width=0.38\columnwidth,trim=0 3mm 0 3mm,clip]{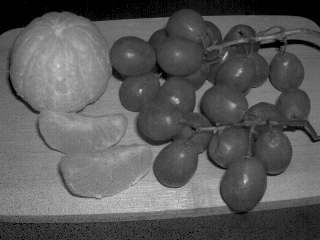}&
 	\includegraphics[width=0.38\columnwidth,trim=0 3mm 0 3mm,clip]{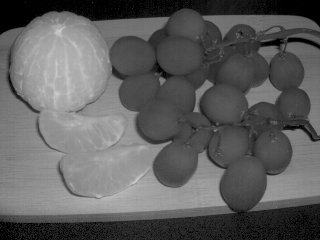}\\[-1mm] (a) Parallel (ground truth) & (b) Crossed (ground truth)	\\[.5mm]
\includegraphics[width=0.38\columnwidth,trim=0 3mm 0 3mm,clip]{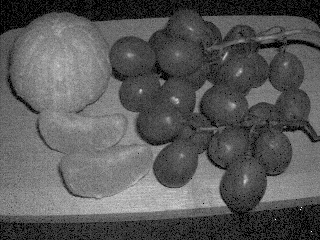}&
	\includegraphics[width=0.38\columnwidth,trim=0 3mm 0 3mm,clip]{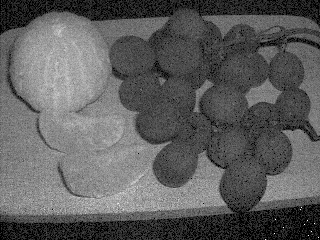}\\[-1mm] (c) Parallel (M1)& (d) Crossed (M1)\\[.5mm]
	\includegraphics[width=0.38\columnwidth,trim=0 3mm 0 3mm,clip]{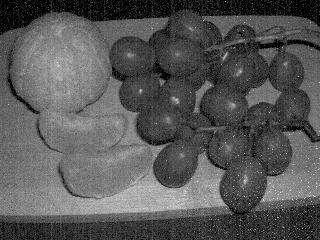}&
	\includegraphics[width=0.38\columnwidth,trim=0 3mm 0 3mm,clip]{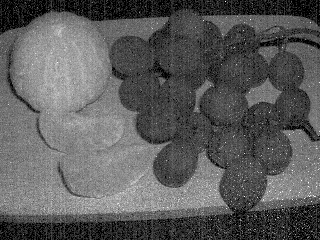}\\[-1mm] (e) Parallel (M1; $\sigma^2_\textrm{read}\!:=\!0$)& (f) Crossed (M1; $\sigma^2_\textrm{read}\!:=\!0$)\\[-1mm]
	\end{tabularx}
	\vspace{-2mm}
	\caption{Source images of the polarization difference image shown in Fig.~\ref{fig:polcompare}a (bottom), reconstructed from the statistics of 400 difference images, with (c,d) and without (e,f) pre-calibrated read noise term. Subfigures (a) and (b) show ground truth images for comparison.}
	\label{fig:pol}
\end{figure}

As another example application of this differential recovery method, Fig.~\ref{fig:pol} shows reconstructions of the source images via M1 from a series of $N\!=\!400$ difference images from the direct-global separation application described in section \ref{sec:direct_global} with and without correction for read noise, as well as ground truth. As expected, polarization-preserving reflections such as specular highlights appear in the ``parallel'' channel only, while sub-surface scattered (depolarized) light contributes to both channels.

Since we exploit basic properties of the Skellam distribution, this method also enables source separation for traditional difference imaging. As this approach essentially enables high-speed spatial multiplexed capture without spatial separation on the sensor, we envision a variety of applications beyond the two presented above.
\section{Discussion}
\label{sec:discussion}

In summary, we propose a new imaging system for direct recording of image differences in a snapshot. The proposed technique directly maps to the emerging technology of time-of-flight sensors and will therefore continue to benefit from the ongoing technological development in that area. The primary benefits of snapshot difference imaging include high video framerates that are only limited by the readout interface as well as lower noise and reduced alignment artifacts as compared to sequential, digital difference imaging. Finally, we devise an algorithm that is capable of encoding and extracting two different images from the mean and variance of a single photograph captured with the proposed method.

\paragraph{Limitations}

Similar to range imaging, most of the demonstrated applications of snapshot difference imaging with time-of-flight sensors require active illumination. Joint coding and precise synchronization between the light sources and the sensor are required. The power of the employed light sources limits the range of distance within which the proposed method would function. 

\paragraph{Future work}

In the future, we would like to explore passive implementations of the proposed method, for example when using them with the natural flicker rates of existing indoor lighting. We would like to explore more sophisticated temporal coding strategies that may be able to separate direct and global illumination based on their temporal characteristics rather than their polarisation properties. We would also like to explore spatio-temporal coding strategies that would allow the light sources to be used as temporally-coded projectors rather than isotropic emitters~\cite{o2015homogeneous}. Finally, we would like to extend the application to mitigating multi-path interference for time-of-flight cameras and other tasks that may benefit from gradient cameras, such as Visual SLAM~\cite{kim2016real} and 3D scanning~\cite{Matsuda:2015}.

\bibliographystyle{ACM-Reference-Format}
\bibliography{references}


\begin{thebibliography}{00}


\ifx \showCODEN    \undefined \def \showCODEN     #1{\unskip}     \fi
\ifx \showDOI      \undefined \def \showDOI       #1{#1}\fi
\ifx \showISBNx    \undefined \def \showISBNx     #1{\unskip}     \fi
\ifx \showISBNxiii \undefined \def \showISBNxiii  #1{\unskip}     \fi
\ifx \showISSN     \undefined \def \showISSN      #1{\unskip}     \fi
\ifx \showLCCN     \undefined \def \showLCCN      #1{\unskip}     \fi
\ifx \shownote     \undefined \def \shownote      #1{#1}          \fi
\ifx \showarticletitle \undefined \def \showarticletitle #1{#1}   \fi
\ifx \showURL      \undefined \def \showURL       {\relax}        \fi
\providecommand\bibfield[2]{#2}
\providecommand\bibinfo[2]{#2}
\providecommand\natexlab[1]{#1}
\providecommand\showeprint[2][]{arXiv:#2}

\bibitem[\protect\citeauthoryear{Bamji, O'Connor, Elkhatib, Mehta, Thompson,
  Prather, Snow, Akkaya, Daniel, Payne, Perry, Fenton, and Chan}{Bamji
  et~al\mbox{.}}{2015}]%
        {Bamji:2015}
\bibfield{author}{\bibinfo{person}{Bamji, C.}, \bibinfo{person}{P. O'Connor},
  \bibinfo{person}{T. Elkhatib}, \bibinfo{person}{S. Mehta},
  \bibinfo{person}{B. Thompson}, \bibinfo{person}{L. Prather},
  \bibinfo{person}{D. Snow}, \bibinfo{person}{O. Akkaya}, \bibinfo{person}{A.
  Daniel}, \bibinfo{person}{A. Payne}, \bibinfo{person}{T. Perry},
  \bibinfo{person}{M. Fenton}, {and} \bibinfo{person}{V.-H. Chan}.}
  \bibinfo{year}{2015}\natexlab{}.
\newblock \showarticletitle{{A 0.13 um CMOS System-on-Chip for a 512 x 424
  Time-of-Flight Image Sensor With Multi-Frequency Photo-Demodulation up to 130
  MHz and 2 GS/s ADC}}.
\newblock \bibinfo{journal}{{\em IEEE Journal of Solid-State Circuits\/}}
  \bibinfo{volume}{50}, \bibinfo{number}{1} (\bibinfo{year}{2015}),
  \bibinfo{pages}{303--319}.
\newblock


\bibitem[\protect\citeauthoryear{Chen, Lensch, Fuchs, and Seidel}{Chen
  et~al\mbox{.}}{2007}]%
        {chen2007polarization}
\bibfield{author}{\bibinfo{person}{Chen, T.}, \bibinfo{person}{H.~P. Lensch},
  \bibinfo{person}{C. Fuchs}, {and} \bibinfo{person}{H.-P. Seidel}.}
  \bibinfo{year}{2007}\natexlab{}.
\newblock \showarticletitle{Polarization and phase-shifting for 3D scanning of
  translucent objects}. In \bibinfo{booktitle}{{\em Proc. IEEE CVPR}}. IEEE,
  \bibinfo{pages}{1--8}.
\newblock


\bibitem[\protect\citeauthoryear{Darmont}{Darmont}{2012}]%
        {darmont2012high}
\bibfield{author}{\bibinfo{person}{Darmont, A.}}
  \bibinfo{year}{2012}\natexlab{}.
\newblock \showarticletitle{High Dynamic Range Imaging: Sensors and
  Architectures}. SPIE.
\newblock


\bibitem[\protect\citeauthoryear{Gottardi, Massari, and Jawed}{Gottardi
  et~al\mbox{.}}{2009}]%
        {gottardi2009100}
\bibfield{author}{\bibinfo{person}{Gottardi, M.}, \bibinfo{person}{N. Massari},
  {and} \bibinfo{person}{S.~A. Jawed}.} \bibinfo{year}{2009}\natexlab{}.
\newblock \showarticletitle{A 100 $\mu$W 128 $\times$ 64 Pixels Contrast-Based
  Asynchronous Binary Vision Sensor for Sensor Networks Applications}.
\newblock \bibinfo{journal}{{\em IEEE Journal of Solid-State Circuits\/}}
  \bibinfo{volume}{44}, \bibinfo{number}{5} (\bibinfo{year}{2009}),
  \bibinfo{pages}{1582--1592}.
\newblock


\bibitem[\protect\citeauthoryear{Hansard, Lee, Choi, and Horaud}{Hansard
  et~al\mbox{.}}{2012}]%
        {Hansard:2012}
\bibfield{author}{\bibinfo{person}{Hansard, M.}, \bibinfo{person}{S. Lee},
  \bibinfo{person}{O. Choi}, {and} \bibinfo{person}{R. Horaud}.}
  \bibinfo{year}{2012}\natexlab{}.
\newblock \bibinfo{booktitle}{{\em Time of Flight Cameras: Principles, Methods,
  and Applications}}.
\newblock \bibinfo{publisher}{Springer}.
\newblock


\bibitem[\protect\citeauthoryear{Heide, Hullin, Gregson, and Heidrich}{Heide
  et~al\mbox{.}}{2013}]%
        {HeideSIG2013}
\bibfield{author}{\bibinfo{person}{Heide, F.}, \bibinfo{person}{M.~B. Hullin},
  \bibinfo{person}{J. Gregson}, {and} \bibinfo{person}{W. Heidrich}.}
  \bibinfo{year}{2013}\natexlab{}.
\newblock \showarticletitle{Low-Budget Transient Imaging using Photonic Mixer
  Devices}.
\newblock \bibinfo{journal}{{\em ACM Trans. Graph. (Proc. SIGGRAPH)\/}}
  \bibinfo{volume}{32}, \bibinfo{number}{4} (\bibinfo{year}{2013}),
  \bibinfo{pages}{45:1--45:10}.
\newblock


\bibitem[\protect\citeauthoryear{Heide, Xiao, Kolb, Hullin, and Heidrich}{Heide
  et~al\mbox{.}}{2014}]%
        {Heide2014scattering}
\bibfield{author}{\bibinfo{person}{Heide, F.}, \bibinfo{person}{L. Xiao},
  \bibinfo{person}{A. Kolb}, \bibinfo{person}{M.~B. Hullin}, {and}
  \bibinfo{person}{W. Heidrich}.} \bibinfo{year}{2014}\natexlab{}.
\newblock \showarticletitle{Imaging in scattering media using correlation image
  sensors and sparse convolutional coding}.
\newblock \bibinfo{journal}{{\em Opt. Express\/}} \bibinfo{volume}{22},
  \bibinfo{number}{21} (\bibinfo{date}{Oct} \bibinfo{year}{2014}),
  \bibinfo{pages}{26338--26350}.
\newblock
\showDOI{%
\url{https://doi.org/10.1364/OE.22.026338}}


\bibitem[\protect\citeauthoryear{Hwang, Kim, and Kweon}{Hwang
  et~al\mbox{.}}{2012}]%
        {hwang2012difference}
\bibfield{author}{\bibinfo{person}{Hwang, Y.}, \bibinfo{person}{J.-S. Kim},
  {and} \bibinfo{person}{I.~S. Kweon}.} \bibinfo{year}{2012}\natexlab{}.
\newblock \showarticletitle{Difference-based image noise modeling using
  {Skellam} distribution}.
\newblock \bibinfo{journal}{{\em IEEE Trans. PAMI\/}} \bibinfo{volume}{34},
  \bibinfo{number}{7} (\bibinfo{year}{2012}), \bibinfo{pages}{1329--1341}.
\newblock


\bibitem[\protect\citeauthoryear{Kadambi, Whyte, Bhandari, Streeter, Barsi,
  Dorrington, and Raskar}{Kadambi et~al\mbox{.}}{2013}]%
        {kadambi2013coded}
\bibfield{author}{\bibinfo{person}{Kadambi, A.}, \bibinfo{person}{R. Whyte},
  \bibinfo{person}{A. Bhandari}, \bibinfo{person}{L. Streeter},
  \bibinfo{person}{C. Barsi}, \bibinfo{person}{A. Dorrington}, {and}
  \bibinfo{person}{R. Raskar}.} \bibinfo{year}{2013}\natexlab{}.
\newblock \showarticletitle{Coded time of flight cameras: sparse deconvolution
  to address multipath interference and recover time profiles}.
\newblock \bibinfo{journal}{{\em ACM Transactions on Graphics (TOG)\/}}
  \bibinfo{volume}{32}, \bibinfo{number}{6} (\bibinfo{year}{2013}),
  \bibinfo{pages}{167}.
\newblock


\bibitem[\protect\citeauthoryear{Kim, Leutenegger, and Davison}{Kim
  et~al\mbox{.}}{2016}]%
        {kim2016real}
\bibfield{author}{\bibinfo{person}{Kim, H.}, \bibinfo{person}{S. Leutenegger},
  {and} \bibinfo{person}{A.~J. Davison}.} \bibinfo{year}{2016}\natexlab{}.
\newblock \showarticletitle{Real-time 3D reconstruction and 6-DoF tracking with
  an event camera}. In \bibinfo{booktitle}{{\em European Conference on Computer
  Vision}}. Springer, \bibinfo{pages}{349--364}.
\newblock


\bibitem[\protect\citeauthoryear{Koppal, Gkioulekas, Young, Park, Crozier,
  Barrows, and Zickler}{Koppal et~al\mbox{.}}{2013}]%
        {Koppal2013TowardWM}
\bibfield{author}{\bibinfo{person}{Koppal, S.~J.}, \bibinfo{person}{I.
  Gkioulekas}, \bibinfo{person}{T. Young}, \bibinfo{person}{H. Park},
  \bibinfo{person}{K.~B. Crozier}, \bibinfo{person}{G.~L. Barrows}, {and}
  \bibinfo{person}{T.~E. Zickler}.} \bibinfo{year}{2013}\natexlab{}.
\newblock \showarticletitle{Toward Wide-Angle Microvision Sensors}.
\newblock \bibinfo{journal}{{\em IEEE Trans. Pattern Anal. Mach. Intell.\/}}
  \bibinfo{volume}{35} (\bibinfo{year}{2013}), \bibinfo{pages}{2982--2996}.
\newblock


\bibitem[\protect\citeauthoryear{Lange, Seitz, Biber, and Schwarte}{Lange
  et~al\mbox{.}}{1999}]%
        {lange1999time}
\bibfield{author}{\bibinfo{person}{Lange, R.}, \bibinfo{person}{P. Seitz},
  \bibinfo{person}{A. Biber}, {and} \bibinfo{person}{R. Schwarte}.}
  \bibinfo{year}{1999}\natexlab{}.
\newblock \showarticletitle{Time-of-flight range imaging with a custom solid
  state image sensor}. In \bibinfo{booktitle}{{\em Industrial Lasers and
  Inspection (EUROPTO Series)}}. International Society for Optics and
  Photonics, \bibinfo{pages}{180--191}.
\newblock


\bibitem[\protect\citeauthoryear{Lichtsteiner, Posch, and
  Delbruck}{Lichtsteiner et~al\mbox{.}}{2008}]%
        {lichtsteiner2008temporalcontrast}
\bibfield{author}{\bibinfo{person}{Lichtsteiner, P.}, \bibinfo{person}{C.
  Posch}, {and} \bibinfo{person}{T. Delbruck}.}
  \bibinfo{year}{2008}\natexlab{}.
\newblock \showarticletitle{A 128$\times$ 128 120 dB 15 $\mu$s latency
  asynchronous temporal contrast vision sensor}.
\newblock \bibinfo{journal}{{\em Solid-State Circuits, IEEE Journal of\/}}
  \bibinfo{volume}{43}, \bibinfo{number}{2} (\bibinfo{year}{2008}),
  \bibinfo{pages}{566--576}.
\newblock


\bibitem[\protect\citeauthoryear{Liu and Gu}{Liu and Gu}{2014}]%
        {liu2014discriminative}
\bibfield{author}{\bibinfo{person}{Liu, C.} {and} \bibinfo{person}{J. Gu}.}
  \bibinfo{year}{2014}\natexlab{}.
\newblock \showarticletitle{Discriminative illumination: Per-pixel
  classification of raw materials based on optimal projections of spectral
  BRDF}.
\newblock \bibinfo{journal}{{\em IEEE Trans. PAMI\/}} \bibinfo{volume}{36},
  \bibinfo{number}{1} (\bibinfo{year}{2014}), \bibinfo{pages}{86--98}.
\newblock


\bibitem[\protect\citeauthoryear{Ma, Hawkins, Peers, Chabert, Weiss, and
  Debevec}{Ma et~al\mbox{.}}{2007}]%
        {ma2007rapid}
\bibfield{author}{\bibinfo{person}{Ma, W.-C.}, \bibinfo{person}{T. Hawkins},
  \bibinfo{person}{P. Peers}, \bibinfo{person}{C.-F. Chabert},
  \bibinfo{person}{M. Weiss}, {and} \bibinfo{person}{P. Debevec}.}
  \bibinfo{year}{2007}\natexlab{}.
\newblock \showarticletitle{Rapid acquisition of specular and diffuse normal
  maps from polarized spherical gradient illumination}. In
  \bibinfo{booktitle}{{\em Proc. EGSR}}. Eurographics Association,
  \bibinfo{pages}{183--194}.
\newblock


\bibitem[\protect\citeauthoryear{Matsuda, Cossairt, and Gupta}{Matsuda
  et~al\mbox{.}}{2015}]%
        {Matsuda:2015}
\bibfield{author}{\bibinfo{person}{Matsuda, N.}, \bibinfo{person}{O. Cossairt},
  {and} \bibinfo{person}{M. Gupta}.} \bibinfo{year}{2015}\natexlab{}.
\newblock \showarticletitle{MC3D: Motion Contrast 3D Scanning}. In
  \bibinfo{booktitle}{{\em Proc. ICCP}}.
\newblock


\bibitem[\protect\citeauthoryear{Nayar and Branzoi}{Nayar and Branzoi}{2003}]%
        {nayar2003adaptive}
\bibfield{author}{\bibinfo{person}{Nayar, S.~K.} {and} \bibinfo{person}{V.
  Branzoi}.} \bibinfo{year}{2003}\natexlab{}.
\newblock \showarticletitle{Adaptive Dynamic Range Imaging: Optical Control of
  Pixel Exposures Over Space and Time.}. In \bibinfo{booktitle}{{\em ICCV}}.
  \bibinfo{pages}{1168--1175}.
\newblock


\bibitem[\protect\citeauthoryear{Nayar, Krishnan, Grossberg, and Raskar}{Nayar
  et~al\mbox{.}}{2006}]%
        {nayar2006fast}
\bibfield{author}{\bibinfo{person}{Nayar, S.~K.}, \bibinfo{person}{G.
  Krishnan}, \bibinfo{person}{M.~D. Grossberg}, {and} \bibinfo{person}{R.
  Raskar}.} \bibinfo{year}{2006}\natexlab{}.
\newblock \showarticletitle{Fast separation of direct and global components of
  a scene using high frequency illumination}.
\newblock \bibinfo{journal}{{\em ACM Trans. Graph. (Proc. SIGGRAPH)\/}}
  \bibinfo{volume}{25}, \bibinfo{number}{3} (\bibinfo{year}{2006}),
  \bibinfo{pages}{935--944}.
\newblock


\bibitem[\protect\citeauthoryear{O'Toole, Achar, Narasimhan, and
  Kutulakos}{O'Toole et~al\mbox{.}}{2015}]%
        {o2015homogeneous}
\bibfield{author}{\bibinfo{person}{O'Toole, M.}, \bibinfo{person}{S. Achar},
  \bibinfo{person}{S.~G. Narasimhan}, {and} \bibinfo{person}{K.~N. Kutulakos}.}
  \bibinfo{year}{2015}\natexlab{}.
\newblock \showarticletitle{Homogeneous codes for energy-efficient illumination
  and imaging}.
\newblock \bibinfo{journal}{{\em ACM Trans. Graph. (Proc. SIGGRAPH)\/}}
  \bibinfo{volume}{34}, \bibinfo{number}{4} (\bibinfo{year}{2015}),
  \bibinfo{pages}{35}.
\newblock


\bibitem[\protect\citeauthoryear{O'Toole, Raskar, and Kutulakos}{O'Toole
  et~al\mbox{.}}{2012}]%
        {otoole2012primal}
\bibfield{author}{\bibinfo{person}{O'Toole, M.}, \bibinfo{person}{R. Raskar},
  {and} \bibinfo{person}{K.~N. Kutulakos}.} \bibinfo{year}{2012}\natexlab{}.
\newblock \showarticletitle{Primal-dual coding to probe light transport}.
\newblock \bibinfo{journal}{{\em ACM Trans. Graph. (Proc. SIGGRAPH)\/}}
  \bibinfo{volume}{31}, \bibinfo{number}{4} (\bibinfo{year}{2012}),
  \bibinfo{pages}{39}.
\newblock


\bibitem[\protect\citeauthoryear{Raskar, Tan, Feris, Yu, and Turk}{Raskar
  et~al\mbox{.}}{2004}]%
        {Raskar:2004}
\bibfield{author}{\bibinfo{person}{Raskar, R.}, \bibinfo{person}{K.-H. Tan},
  \bibinfo{person}{R. Feris}, \bibinfo{person}{J. Yu}, {and}
  \bibinfo{person}{M. Turk}.} \bibinfo{year}{2004}\natexlab{}.
\newblock \showarticletitle{Non-photorealistic Camera: Depth Edge Detection and
  Stylized Rendering Using Multi-flash Imaging}.
\newblock \bibinfo{journal}{{\em ACM Trans. Graph. (Proc. SIGGRAPH)\/}}
  \bibinfo{volume}{23}, \bibinfo{number}{3} (\bibinfo{year}{2004}),
  \bibinfo{pages}{679--688}.
\newblock


\bibitem[\protect\citeauthoryear{Schmidt}{Schmidt}{2011}]%
        {schmidt2011analysis}
\bibfield{author}{\bibinfo{person}{Schmidt, M.}}
  \bibinfo{year}{2011}\natexlab{}.
\newblock {\em \bibinfo{title}{Analysis, modeling and dynamic optimization of
  {3D} time-of-flight imaging systems}}.
\newblock \bibinfo{thesistype}{Ph.D. Dissertation}. \bibinfo{school}{University
  of Heidelberg}.
\newblock


\bibitem[\protect\citeauthoryear{Shrestha, Heide, Heidrich, and
  Wetzstein}{Shrestha et~al\mbox{.}}{2016}]%
        {Shrestha:2016}
\bibfield{author}{\bibinfo{person}{Shrestha, S.}, \bibinfo{person}{F. Heide},
  \bibinfo{person}{W. Heidrich}, {and} \bibinfo{person}{G. Wetzstein}.}
  \bibinfo{year}{2016}\natexlab{}.
\newblock \showarticletitle{Computational Imaging with Multi-Camera
  Time-of-Flight Systems}.
\newblock \bibinfo{journal}{{\em ACM Trans. Graph. (Proc. SIGGRAPH)\/}}
  (\bibinfo{year}{2016}).
\newblock


\bibitem[\protect\citeauthoryear{Skellam}{Skellam}{1946}]%
        {skellam}
\bibfield{author}{\bibinfo{person}{Skellam, J.~G.}}
  \bibinfo{year}{1946}\natexlab{}.
\newblock \showarticletitle{The Frequency Distribution of the Difference
  Between Two Poisson Variates Belonging to Different Populations}.
\newblock \bibinfo{journal}{{\em Journal of the Royal Statistical Society\/}}
  \bibinfo{volume}{109}, \bibinfo{number}{3} (\bibinfo{year}{1946}),
  \bibinfo{pages}{296--296}.
\newblock
\showISSN{09528385}


\bibitem[\protect\citeauthoryear{Solhusvik, Kuang, Lin, Manabe, Lyu, Rhodes,
  et~al\mbox{.}}{Solhusvik et~al\mbox{.}}{2013}]%
        {solhusvik2013comparison}
\bibfield{author}{\bibinfo{person}{Solhusvik, J.}, \bibinfo{person}{J. Kuang},
  \bibinfo{person}{Z. Lin}, \bibinfo{person}{S. Manabe}, \bibinfo{person}{J.
  Lyu}, \bibinfo{person}{H. Rhodes}, {et~al\mbox{.}}}
  \bibinfo{year}{2013}\natexlab{}.
\newblock \showarticletitle{A comparison of high dynamic range CIS technologies
  for automotive applications}. In \bibinfo{booktitle}{{\em Proc. 2013 Int.
  Image Sensor Workshop (IISW)}}.
\newblock


\bibitem[\protect\citeauthoryear{Tadano, Pediredla, and Veeraraghavan}{Tadano
  et~al\mbox{.}}{2015}]%
        {Tadano2015}
\bibfield{author}{\bibinfo{person}{Tadano, R.}, \bibinfo{person}{A.~K.
  Pediredla}, {and} \bibinfo{person}{A. Veeraraghavan}.}
  \bibinfo{year}{2015}\natexlab{}.
\newblock \showarticletitle{Depth Selective Camera: A Direct, On-Chip,
  Programmable Technique for Depth Selectivity in Photography}. In
  \bibinfo{booktitle}{{\em 2015 {IEEE} International Conference on Computer
  Vision ({ICCV})}}. \bibinfo{publisher}{{IEEE}}.
\newblock


\bibitem[\protect\citeauthoryear{Tomasi and Manduchi}{Tomasi and
  Manduchi}{1998}]%
        {tomasi1998bilateral}
\bibfield{author}{\bibinfo{person}{Tomasi, C.} {and} \bibinfo{person}{R.
  Manduchi}.} \bibinfo{year}{1998}\natexlab{}.
\newblock \showarticletitle{Bilateral filtering for gray and color images}. In
  \bibinfo{booktitle}{{\em Proc. IEEE ICCV}}. IEEE, \bibinfo{pages}{839--846}.
\newblock


\bibitem[\protect\citeauthoryear{Tumblin, Agrawal, and Raskar}{Tumblin
  et~al\mbox{.}}{2005}]%
        {tumblin2005gradient}
\bibfield{author}{\bibinfo{person}{Tumblin, J.}, \bibinfo{person}{A. Agrawal},
  {and} \bibinfo{person}{R. Raskar}.} \bibinfo{year}{2005}\natexlab{}.
\newblock \showarticletitle{Why {I} want a gradient camera}. In
  \bibinfo{booktitle}{{\em Proc. IEEE CVPR}}, Vol.~\bibinfo{volume}{1}.
  \bibinfo{pages}{103--110}.
\newblock


\bibitem[\protect\citeauthoryear{Wan, Li, Agranov, Levoy, and Horowitz}{Wan
  et~al\mbox{.}}{2012}]%
        {wan2012cmos}
\bibfield{author}{\bibinfo{person}{Wan, G.}, \bibinfo{person}{X. Li},
  \bibinfo{person}{G. Agranov}, \bibinfo{person}{M. Levoy}, {and}
  \bibinfo{person}{M. Horowitz}.} \bibinfo{year}{2012}\natexlab{}.
\newblock \showarticletitle{CMOS image sensors with multi-bucket pixels for
  computational photography}.
\newblock \bibinfo{journal}{{\em Solid-State Circuits, IEEE Journal of\/}}
  \bibinfo{volume}{47}, \bibinfo{number}{4} (\bibinfo{year}{2012}),
  \bibinfo{pages}{1031--1042}.
\newblock


\bibitem[\protect\citeauthoryear{Wang and Molnar}{Wang and Molnar}{2012}]%
        {Wang2012}
\bibfield{author}{\bibinfo{person}{Wang, A.} {and} \bibinfo{person}{A.
  Molnar}.} \bibinfo{year}{2012}\natexlab{}.
\newblock \showarticletitle{A Light-Field Image Sensor in 180 nm {CMOS}}.
\newblock \bibinfo{journal}{{\em {IEEE} Journal of Solid-State Circuits\/}}
  \bibinfo{volume}{47}, \bibinfo{number}{1} (\bibinfo{date}{jan}
  \bibinfo{year}{2012}), \bibinfo{pages}{257--271}.
\newblock


\bibitem[\protect\citeauthoryear{Wang, Sivaramakrishnan, and Molnar}{Wang
  et~al\mbox{.}}{2012}]%
        {wang2012180nm}
\bibfield{author}{\bibinfo{person}{Wang, A.}, \bibinfo{person}{S.
  Sivaramakrishnan}, {and} \bibinfo{person}{A. Molnar}.}
  \bibinfo{year}{2012}\natexlab{}.
\newblock \showarticletitle{A 180nm cmos image sensor with on-chip
  optoelectronic image compression}. In \bibinfo{booktitle}{{\em Custom
  Integrated Circuits Conference (CICC), 2012 IEEE}}. IEEE,
  \bibinfo{pages}{1--4}.
\newblock


\bibitem[\protect\citeauthoryear{Weikersdorfer, Adrian, Cremers, and
  Conradt}{Weikersdorfer et~al\mbox{.}}{2014}]%
        {weikersdorfer2014event}
\bibfield{author}{\bibinfo{person}{Weikersdorfer, D.}, \bibinfo{person}{D.~B.
  Adrian}, \bibinfo{person}{D. Cremers}, {and} \bibinfo{person}{J. Conradt}.}
  \bibinfo{year}{2014}\natexlab{}.
\newblock \showarticletitle{Event-based 3D SLAM with a depth-augmented dynamic
  vision sensor}. In \bibinfo{booktitle}{{\em Robotics and Automation (ICRA),
  2014 IEEE International Conference on}}. IEEE, \bibinfo{pages}{359--364}.
\newblock


\bibitem[\protect\citeauthoryear{Willassen, Solhusvik, Johansson, Yaghmai,
  Rhodes, Manabe, Mao, Lin, Yang, Cellek, et~al\mbox{.}}{Willassen
  et~al\mbox{.}}{}]%
        {willassen20151280}
\bibfield{author}{\bibinfo{person}{Willassen, T.}, \bibinfo{person}{J.
  Solhusvik}, \bibinfo{person}{R. Johansson}, \bibinfo{person}{S. Yaghmai},
  \bibinfo{person}{H. Rhodes}, \bibinfo{person}{S. Manabe}, \bibinfo{person}{D.
  Mao}, \bibinfo{person}{Z. Lin}, \bibinfo{person}{D. Yang},
  \bibinfo{person}{O. Cellek}, {et~al\mbox{.}}}
\newblock \showarticletitle{A 1280$\times$ 1080 4.2 $\mu$m split-diode pixel
  HDR sensor in 110nm BSI CMOS process}.
\newblock


\bibitem[\protect\citeauthoryear{Wolff}{Wolff}{1990}]%
        {wolff1990polarization}
\bibfield{author}{\bibinfo{person}{Wolff, L.~B.}}
  \bibinfo{year}{1990}\natexlab{}.
\newblock \showarticletitle{Polarization-based material classification from
  specular reflection}.
\newblock \bibinfo{journal}{{\em IEEE Transactions on Pattern Analysis and
  Machine Intelligence\/}} \bibinfo{volume}{12}, \bibinfo{number}{11}
  (\bibinfo{year}{1990}).
\newblock


\bibitem[\protect\citeauthoryear{Woodham}{Woodham}{1980}]%
        {woodham1980photometric}
\bibfield{author}{\bibinfo{person}{Woodham, R.~J.}}
  \bibinfo{year}{1980}\natexlab{}.
\newblock \showarticletitle{Photometric method for determining surface
  orientation from multiple images}.
\newblock \bibinfo{journal}{{\em Optical engineering\/}} \bibinfo{volume}{19},
  \bibinfo{number}{1} (\bibinfo{year}{1980}), \bibinfo{pages}{191139--191139}.
\newblock


\bibitem[\protect\citeauthoryear{Zomet and Nayar}{Zomet and Nayar}{2006}]%
        {zomet2006lensless}
\bibfield{author}{\bibinfo{person}{Zomet, A.} {and} \bibinfo{person}{S.~K.
  Nayar}.} \bibinfo{year}{2006}\natexlab{}.
\newblock \showarticletitle{Lensless imaging with a controllable aperture}. In
  \bibinfo{booktitle}{{\em Computer Vision and Pattern Recognition, 2006 IEEE
  Computer Society Conference on}}, Vol.~\bibinfo{volume}{1}. IEEE,
  \bibinfo{pages}{339--346}.
\newblock


\end{thebibliography}

\appendix
\section{Minimal setup}\label{sec:appendix}

\begin{minipage}{0.66\columnwidth}
A basic implementation of a snapshot difference imager for infrared only can be obtained by connecting a TI OPT8241-CDK sensor board with two of its original infrared light sources via an extended ribbon cable (Fig.\ref{fig:cablemod}). The modulation polarity is reversed (wires \#10 and \#12 swapped) and wire \#16 is cut for the second light source. Note, however, that the OPT8241-CDK board by itself cannot generate modulation signals below 10\,MHz. Since the camera captures groups of four phase-shifted sub-frames (in 90$^\circ$ steps), the effective frame rate is reduced by a factor of 4 compared to our system. \end{minipage}%
\hfill%
\begin{minipage}{0.3\columnwidth}

\begin{center}
    \vspace{-8mm}\includegraphics[angle=-90,width=0.8\linewidth]{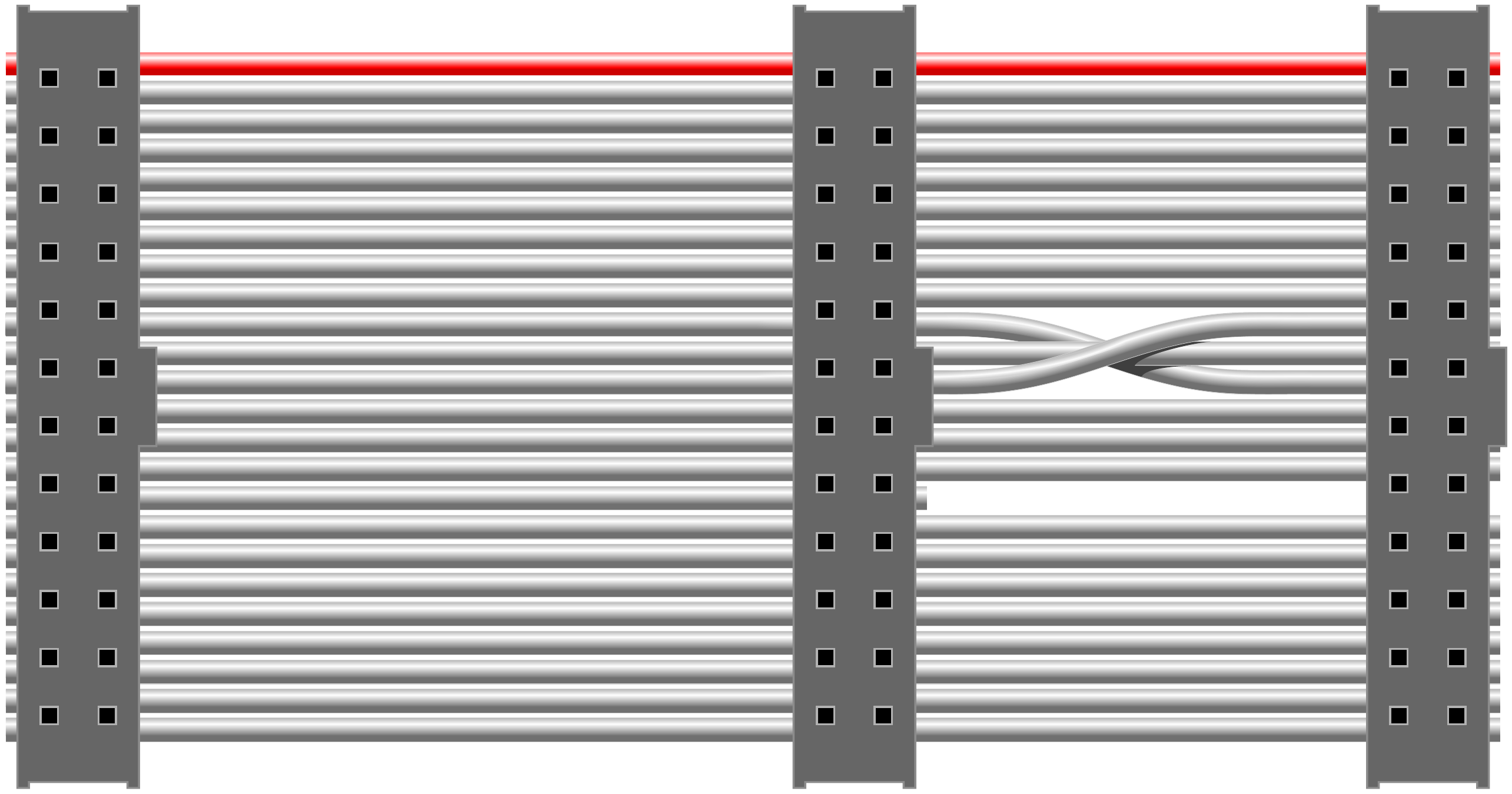}\vspace{-0.1mm}%
    \captionof{figure}{Modified ribbon cable with (top to bottom) connectors for sensor board, LS1 and LS2.}
    \label{fig:cablemod}
\end{center}
\end{minipage}


\end{document}